\documentclass{article}

\PassOptionsToPackage{dvipsnames}{xcolor}

\usepackage{microtype}
\usepackage{graphicx}
\usepackage{subfigure}
\usepackage{booktabs} 
\usepackage{upgreek}
\usepackage[accepted]{icml2024}
\usepackage{caption}
\usepackage{colortbl}

\usepackage{amsmath}
\usepackage{amssymb}
\usepackage{multirow}

\usepackage{array}
\newcolumntype{P}[1]{>{\centering\arraybackslash}p{#1}}

\usepackage{multicol}
\usepackage{multirow}
\usepackage{changepage}
\usepackage{xr}
\usepackage{xr-hyper}
\usepackage{hyperref}

\graphicspath{ {./images/} }

\usepackage[capitalize,noabbrev]{cleveref}
\crefname{section}{Sec.}{Secs.}
\Crefname{section}{Section}{Sections}
\Crefname{table}{Table}{Tables}
\crefname{table}{Tab.}{Tabs.}

\hypersetup{
	hidelinks,
	colorlinks=true,
	linkcolor=Blue,
	filecolor=Blue,
	citecolor=Blue,
	urlcolor=Blue,
	breaklinks=false
}

\newcommand{\myTitle}{Slicedit: Zero-Shot Video Editing With Text-to-Image Diffusion Models Using Spatio-Temporal Slices}

\makeatletter
\newcommand*{\addFileDependency}[1]{
	\typeout{(#1)}
	\@addtofilelist{#1}
	\IfFileExists{#1}{\typeout{File #1 O.K.}}{\typeout{No file #1.}}
}
\makeatother

\newcommand{\eg}{\emph{e.g.\@ }}
\newcommand{\ie}{\emph{i.e.\@ }}

\icmltitlerunning{\myTitle}

\begin{document}
\twocolumn[{
\icmltitle{\myTitle}



\icmlsetsymbol{equal}{*}

\begin{icmlauthorlist}
\icmlauthor{Nathaniel Cohen}{equal,xxx,yyy}
\icmlauthor{Vladimir Kulikov}{equal,yyy}
\icmlauthor{Matan Kleiner}{equal,yyy}
\icmlauthor{Inbar Huberman-Spiegelglas}{yyy}
\icmlauthor{Tomer Michaeli}{yyy}
\end{icmlauthorlist}

\icmlaffiliation{yyy}{Technion -- Israel Institute of Technology, Haifa, Israel}
\icmlaffiliation{xxx}{Mines Paris -- PSL Research University, Paris, France}

\icmlcorrespondingauthor{Vladimir Kulikov}{vladimir.k@campus.technion.ac.il}

\icmlkeywords{Machine Learning, ICML, Video, Video Editing, Text-guided Video Editing, Text-to-Video, Text-to-Image}

\vskip -0.75cm

\begin{center}
\centering
\captionsetup{type=figure}
\includegraphics[width=\textwidth]{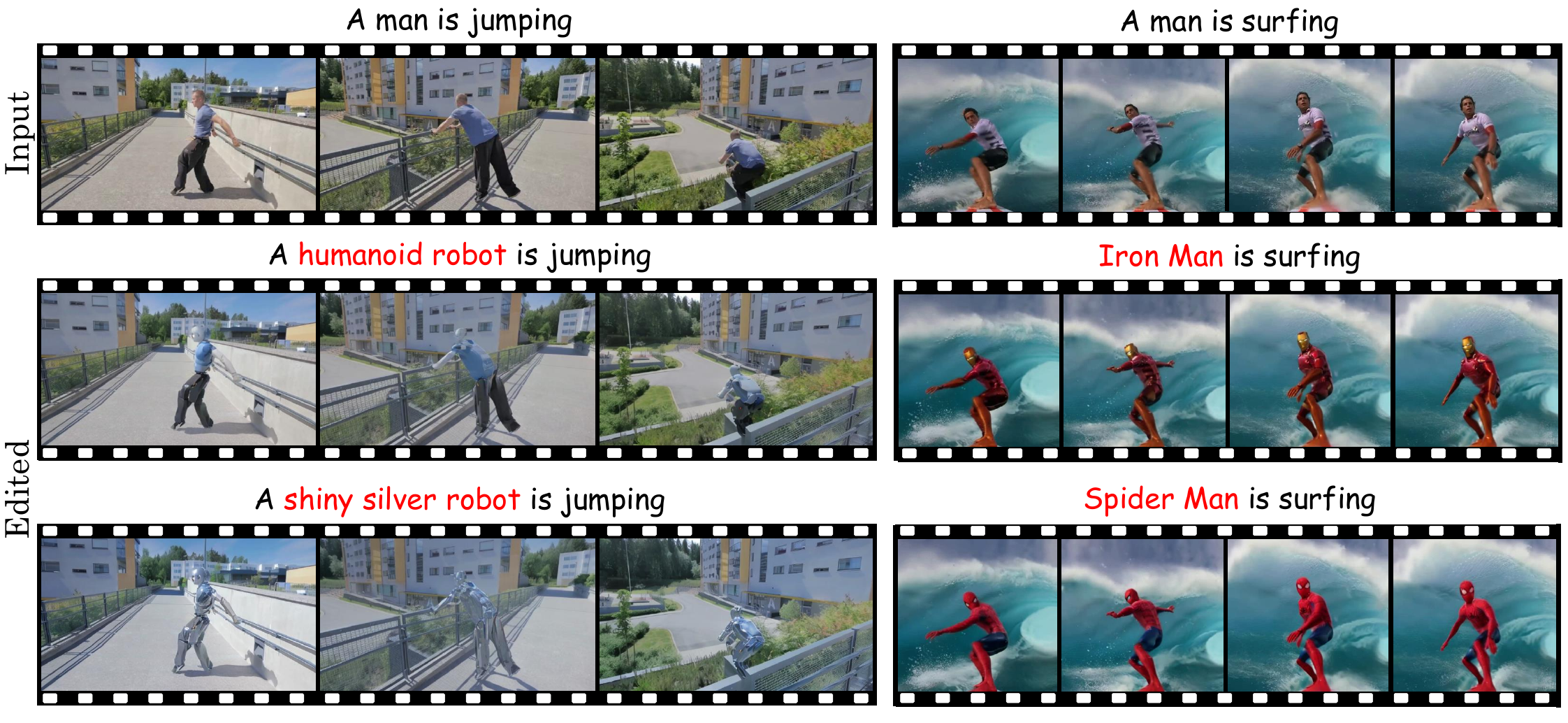}
\caption{\textbf{Slicedit.} We present a zero-shot method for text-based video editing based on a pretrained text-to-image diffusion model. 
Our method can edit challenging long videos with complex nonrigid motion and occlusions, while preserving the regions not specified in the text prompt (\textit{e.g.}~changing only the person into a robot without affecting the background). See our
\href{https://matankleiner.github.io/slicedit/}{website} for video results.}
\label{fig:teaser}
\end{center}

}
]


\printAffiliationsAndNotice{\icmlEqualContribution} 

\begin{abstract}
Text-to-image (T2I) diffusion models achieve state-of-the-art results in image synthesis and editing. However, leveraging such pretrained models for video editing is considered a major challenge. 
Many existing works attempt to enforce temporal consistency in the edited video through explicit correspondence mechanisms, either in pixel space or between deep features. 
These methods, however, struggle with strong nonrigid motion.
In this paper, we introduce a fundamentally different approach, which is based on the observation that spatiotemporal slices of natural videos exhibit similar characteristics to natural images. Thus, the same T2I diffusion model that is normally used only as a prior on video frames, can also serve as a strong prior for enhancing temporal consistency by applying it on spatiotemporal slices. Based on this observation, we present Slicedit, a method for text-based video editing that utilizes a pretrained T2I diffusion model to process both spatial and spatiotemporal slices. 
Our method generates videos that retain the structure and motion of the original video while adhering to the target text. 
Through extensive experiments, we demonstrate Slicedit's ability to edit a wide range of real-world videos, confirming its clear advantages compared to existing competing methods. 
\end{abstract}

\section{Introduction}

\label{sec:intro}
Text-to-image (T2I) diffusion models have reached remarkable capabilities, enabling high-quality image synthesis that can be controlled by highly descriptive text prompts~\citep{Rombach22LDM, saharia2022photorealistic, dai2023emu, betker2023improving}. These capabilities have been shown to enable text-based editing of real images~\citep{Hertz22p2p, Bahjat22Imagic, Brooks22InstructPix2Pix, Tumanyan2023PnP, hubermanspiegelglas23, zhang2023adding}. However, using T2I models in a zero-shot manner for \emph{video editing} is still considered an open challenge, especially when it comes to long videos with strong nonrigid motion and occlusions. 

The naive approach of using T2I models for editing a video frame-by-frame leads to temporal inconsistencies~\citep{Wu23TuneAVideo}, both over short periods of time (\eg flickering) and over long durations (\eg drift in appearance). 
To mitigate temporal inconsistencies, previous and concurrent video editing methods use extended attention, which enables editing multiple frames jointly and hence improves temporal consistency~\citep{Wu23TuneAVideo, QI2023FateZero, Ceylan2023Pix2Video, Khachatryan2023text2video-zero}. However, relying solely on extended attention often results in inconsistent editing of textures and fine details. Other concurrent methods tackle temporal inconsistency by using feature correspondence across frames~\citep{Yang23Rerender, geyer2024tokenflow}. These methods tend to fail in scenarios where the correspondences are weak, \eg in long videos or in videos with fast and complex nonrigid motion.

Here we introduce a new approach, which we coin Slicedit\footnote{Slicedit can be pronounced ``slice-edit'' or ``sliced-it''.}, to edit videos in a zero-shot manner. Similarly to recent and concurrent works, our method ``inflates'' a pretrained T2I latent diffusion model into a model suitable for video editing, doing so in a zero-shot manner (\ie without fine-tuning the model). 
To edit a real video, we use the DDPM inversion method of \citet{hubermanspiegelglas23} with our inflated denoiser to obtain the sequence of noise vectors that causes the diffusion process to generate that video. We then re-generate the video with the text prompt provided by the user, while fixing the noise vectors and injecting features obtained from the inversion of the source video, as previously shown to be useful by~\citet{QI2023FateZero}.

\begin{figure*}
\centering
\includegraphics[width=0.95\textwidth]{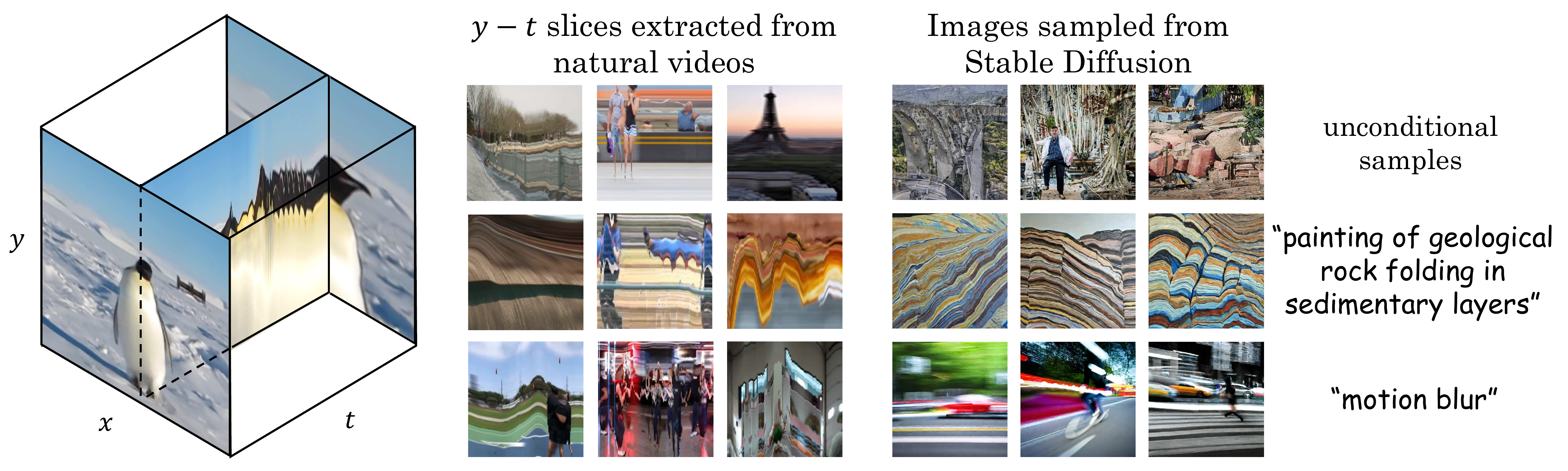}
\caption{\textbf{Diffusion model as a spatiotemporal slice prior.} The left pane shows the $(\mathbf{x},\mathbf{y},t)$ space-time volume of a video. The middle pane shows $\mathbf{y}-t$ slices of natural videos. The right pane shows images generated by Stable Diffusion using several text-prompts. The generated images have similar characteristics to spatiotemporal slices of natural videos. This suggests that a pretrained text-to-image model can serve as a good prior for spatiotemporal video slices.}
\label{fig:temporal_slices}
\end{figure*}

Our inflated denoiser deviates from the pretrained T2I model in two fundamental aspects: (i) Similarly to concurrent works, the self-attention modules are converted to extended attention, capturing the dynamics between frames. (ii) More importantly, in addition to denoising frames, we enforce temporal consistency by also processing \emph{spatiotemporal slices} of the video volume. Specifically, we observe that spatiotemporal slices share similarities with natural images (Fig.~\ref{fig:temporal_slices}), prompting us to leverage the same pretrained T2I denoiser for their denoising. Thus, we apply the T2I model on $\mathbf{x}-t$ or $\mathbf{y}-t$ slices within the $(\mathbf{x},\mathbf{y},t)$ space-time volume (left pane of Fig.~\ref{fig:temporal_slices}) 
and merge the result with the denoised frames. This multi-axis denoising is key to our method's ability to alter only specified regions in a video while keeping the remaining content fixed.

As we show, our method can edit videos while maintaining temporal consistency, even in cases with strong nonrigid motion or occlusions. As illustrated in Fig.~\ref{fig:teaser}, our method preserves video structure and layout in unedited regions, \eg editing only the man while keeping his surroundings unaltered. We evaluate Slicedit on real-world videos, both short and long, and across a diverse range of motion types. Compared with state-of-the-art zero-shot video editing methods, our results demonstrate a clear advantage for utilizing the untapped potential of spatiotemporal slices.

\section{Related work}
\label{sec:related}

\paragraph{Tuning T2I models for video editing.} A common approach for harnessing pretrained T2I models for video editing, is to introduce architectural changes to the model and then tune it on videos~\citep{Wu23TuneAVideo, liu2023video-p2p, Wang2023vid2vid-zero, xing2023simda, esser2023structure, feng2023ccedit}.   
The architectural modifications are usually in the form of insertion of (spatio) temporal layers, like depth-wise 3D convolution, extended attention across frames, and 1D self-attention across time. Those modifications enhance the model's ability to maintain temporal consistency. 
The tuning process usually starts from the weights of the pretrained T2I model, where in some cases only the weights of the new layers are tuned, and uses a large dataset of text-video pairs. This results in a computationally intensive and time consuming training process. Alternatively, the tuning can be done on a single video (the one being edited), yet this approach is time consuming at inference, as each video requires its own tuning process.

\paragraph{T2I models for zero-shot video editing.} Some methods use T2I models for video editing without any parameter tuning, in a zero-shot manner. Our work falls under this category. Similarly to methods that require fine-tuning, most zero-shot methods adjust the attention modules of the model to capture associations across time~\citep{Khachatryan2023text2video-zero, QI2023FateZero, Ceylan2023Pix2Video, Yang23Rerender, geyer2024tokenflow, cong2024flatten, zhang2024controlvideo}. Relying on extended attention for temporal consistency, as done in~\citep{Khachatryan2023text2video-zero, QI2023FateZero}, results in inconsistent editing of textures and fine details. Thus, some methods~\citep{Ceylan2023Pix2Video, Yang23Rerender, zhang2024controlvideo} also use other mechanisms, like conditioning on depth or edge maps, usually using a pretrained ControlNet~\citep{zhang2023adding}. This improves the quality and temporal consistency of the edited video, but introduces artifacts when the control signal is partial or inaccurate. Another method, used by~\citet{Yang23Rerender, geyer2024tokenflow}, is to edit only key frames and propagate their features to all other video frames. This method strongly enforces temporal consistency, however, it breaks under fast motion, severe occlusions, and when the correspondences weaken over time. Our method also uses extended attention, however we also use spatiotemporal slices as a way of enforcing temporal consistency. 

\paragraph{Spatiotemporal slices of the space-time volume.} 
The $\mathbf{x}-t$ or $\mathbf{y}-t$ slices of space-time video volumes have been used for various different tasks, 
including energy models of motion perception~\citep{adelson1985spatiotemporal}, epipolar plane image analysis~\citep{bolles1987epipolar}, motion analysis~\citep{ngo2003motion}, video mosaics~\citep{rav2005dynamosaics}, video quality assessment~\citep{vu2014vis} and temporal super resolution~\citep{zuckerman2020across}. In particular,~\citet{rav2005dynamosaics} used the spatiotemporal slices of a video to generate new videos, leveraging their similarity to natural images. In addition,~\citet{zuckerman2020across} observed that small patches in natural videos are similar across both the spatial and spatiotemporal dimensions of the space-time volume. Our method also leverages the similarity between spatiotemporal slices and natural images, in that we use an image prior to regularize spatiotemporal slices so as to enforce temporal consistency.

\section{Preliminaries}
\label{preliminaries}

Denoising Diffusion Probabilistic Models (DDPMs)~\citep{ho2020denoising} are a class of generative models that aim to approximate a data distribution through a progressive denoising process. Generating a sample $x_0$ from those models starts by randomly drawing Gaussian noise, $x_T \sim \mathcal{N} (0, \mathbf{I})$, and recursively denoising it in $T$ denoising steps.
The denoiser $\epsilon_\theta(\cdot)$ is usually a U-Net~\citep{ronneberger2015unet}, trained to predict the noise at each timestep. 
This denoiser can be conditioned on various types of signals, usually via a cross-attention module. Here we use models that are conditioned on a text prompt, $p$, yielding $\epsilon_\theta(\cdot,p)$.    
The diffusion process can be applied in pixel space or in the latent space of some encoder, which results in a latent diffusion model~\cite{Rombach22LDM}. In the latter case, the generated samples are decoded back to pixel space by a decoder.
 
A pretrained DDPM can be used not only for generating synthetic images, but also for editing real images. A common approach to do so involves an inversion process, which extracts noise vectors that generate the given image when used in the sampling process. Many methods focus on inversion for the DDIM sampling scheme. \citet{hubermanspiegelglas23} introduced an editing approach that is based on inversion of the DDPM scheme. This method has been illustrated to more strongly preserve the structure of the source image than DDIM inversion, and thus to lead to better editing results. The method of  \citet{hubermanspiegelglas23}  starts by generating increasingly noisy versions of the input image, one for each diffusion timestep. Subsequently, the noise vector for each timestep $\tau$ of the generation process is extracted from $x_{\tau-1}$ and $\epsilon_\theta(x_{\tau},p)$ (see App.~\ref{sm:algorithmDDPM} for more details). During the inversion process, the T2I denoiser can be optionally conditioned on a source prompt, $p_{\text{src}}$, describing the input image.
After the noise vectors have been extracted, they are used in the DDPM generation process, while conditioning the denoiser on a target prompt, $p_{\text{tar}}$, describing the desired output. The edited image preserves fidelity to the original image while adhering to the new prompt.

\section{Method}
\label{sec:method}

\begin{figure}[t]
\centering
\includegraphics[width=0.9\columnwidth]{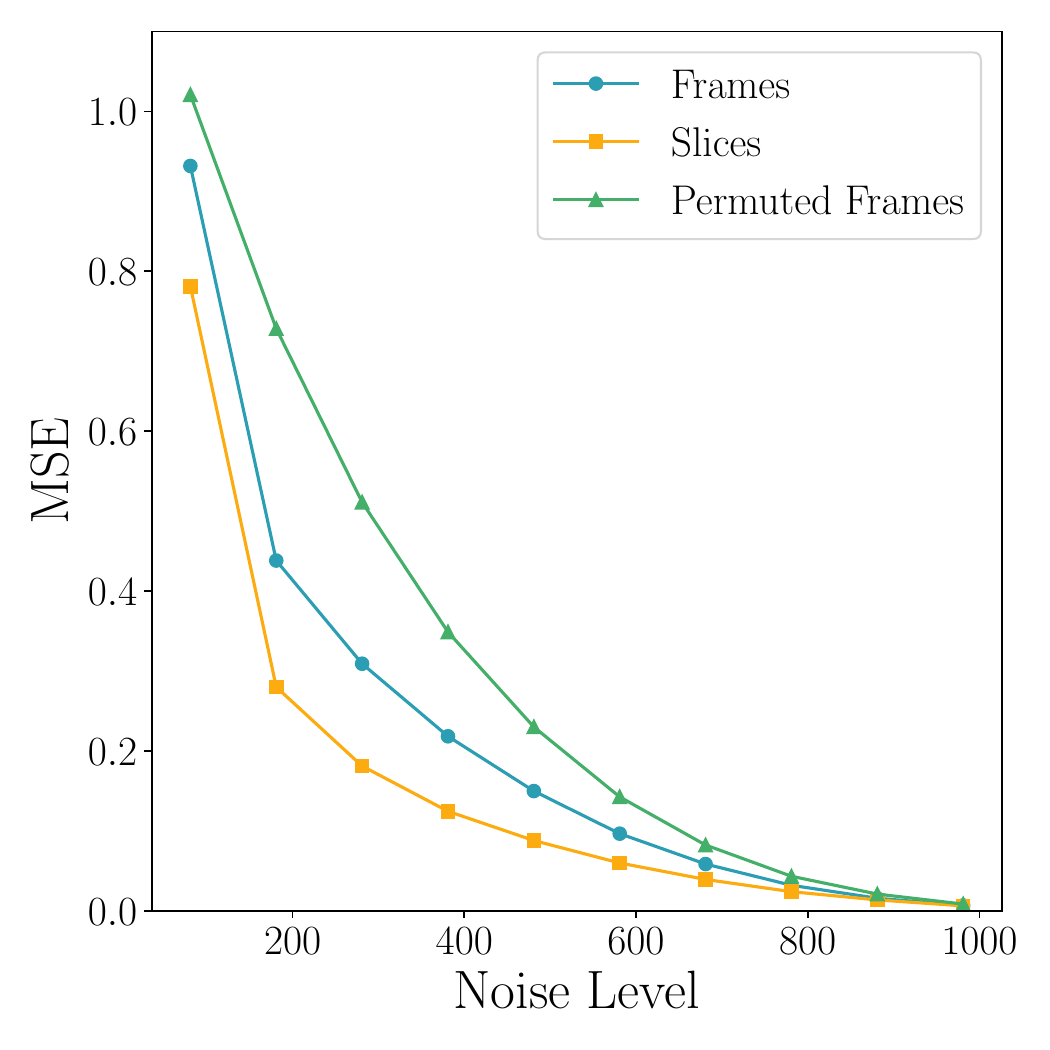}
\caption{\textbf{Applying a pretrained image denoiser to spatiotemporal slices.} The plot shows the MSE obtained when applying a pretrained Stable Diffusion denoiser to noisy video frames, spatiotemporal slices and permuted frames (all in latent space). The MSE for spatiotemporal slices is comparable to (even lower than) the MSE for frames, and both are lower than the MSE for permuted frames, which are out-of-distribution for the denoiser.}
\label{fig:MSE_sd_denoiser}
\end{figure}

Our method requires the following inputs: a video volume $I_0$, a source prompt $p_{\text{src}}$ describing the video, and a target prompt $p_{\text{tar}}$ describing the desired edited video. The goal is to generate an edited video $J_0$ that adheres to $p_{\text{tar}}$ while preserving the original motion and layout of $I_0$. As depicted in Fig.~\ref{fig:overview}, the axes of the video space-time volume are denoted by $(\mathbf{x},\mathbf{y},t)$, where $\mathbf{x}-\mathbf{y}$ planes correspond to video frames and $\mathbf{y}-t$ planes are referred to as spatiotemporal slices.

\subsection{Inflated Denoiser}

We inflate the T2I denoiser $\epsilon_\theta(\cdot,p)$ into a video denoiser $\epsilon^\text{V}_\theta(\cdot,p)$ by performing the following modifications.

\paragraph{Extended attention.} The original denoiser, $\epsilon_\theta(\cdot,p)$, is a U-Net, comprised of residual, cross-attention and self-attention blocks~\citep{Rombach22LDM}. Following the general approach proposed by~\citet{Wu23TuneAVideo} and used by others \cite{geyer2024tokenflow}, we modify the self-attention within each transformer block into an extended attention module. 
This enables the attention module to process multiple frames together, resulting in an attention map with correspondence between multiple frames. Our extended attention is calculated between each frame and a set of 3 key-frames, consisting of a global frame strategically positioned at half of the video length, along with two local frames which are chosen as the $2^\text{nd}$ and $5^\text{th}$ from within the 6-frame processing window containing the target frame. The global key-frame facilitate collaborative editing across the entire video, while the local key-frames help preserve temporal consistency among closely situated frames. The new denoiser, denoted $\epsilon^{\text{EA}}_\theta(\cdot,p)$, does not require any further training, similarly to previous works~\cite{QI2023FateZero, Ceylan2023Pix2Video, geyer2024tokenflow}. 
More details can be found in App.~\ref{sm:ext_attent} and an illustartion can be found in Fig.~\ref{fig:SM_ext_attn}.  

\begin{figure*}
\centering
\includegraphics[width=\textwidth]{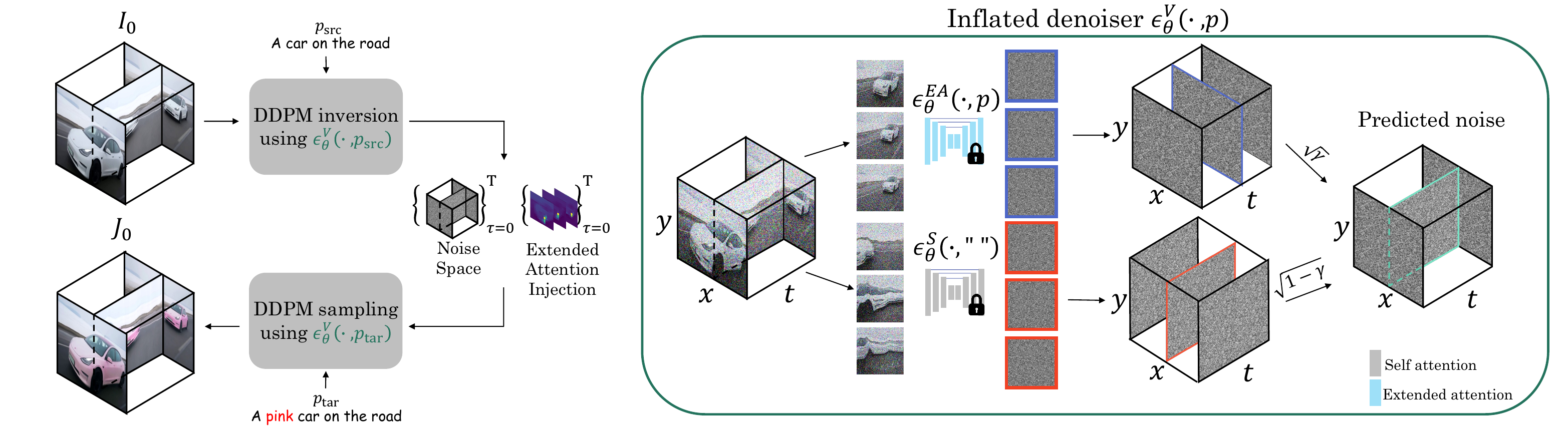}
\caption{\textbf{Slicedit overview.} Left: To edit a video $I_0$, we apply DDPM inversion using our video-denoising model, which is an inflated version of the T2I model. This process extracts noise volumes and attention maps for each diffusion timestep. Subsequently, we run DDPM sampling using the extracted noise space, while injecting the extended attention maps at specific timesteps. The inversion and sampling are performed while conditioning the inflated denoiser on the source and target text prompts, respectively. Right: Our inflated denoiser employs two versions of the pretrained image denoiser. A version with extended attention is applied to $\mathbf{x}-\mathbf{y}$ slices (blue), and the original denoiser is applied to $\mathbf{y}-t$ slices (red). The two predicted noise volumes are then combined into the final predicted noise volume (marked in green).}
\label{fig:overview}
\end{figure*}

\begin{figure*}[t]
\includegraphics[width=\textwidth]{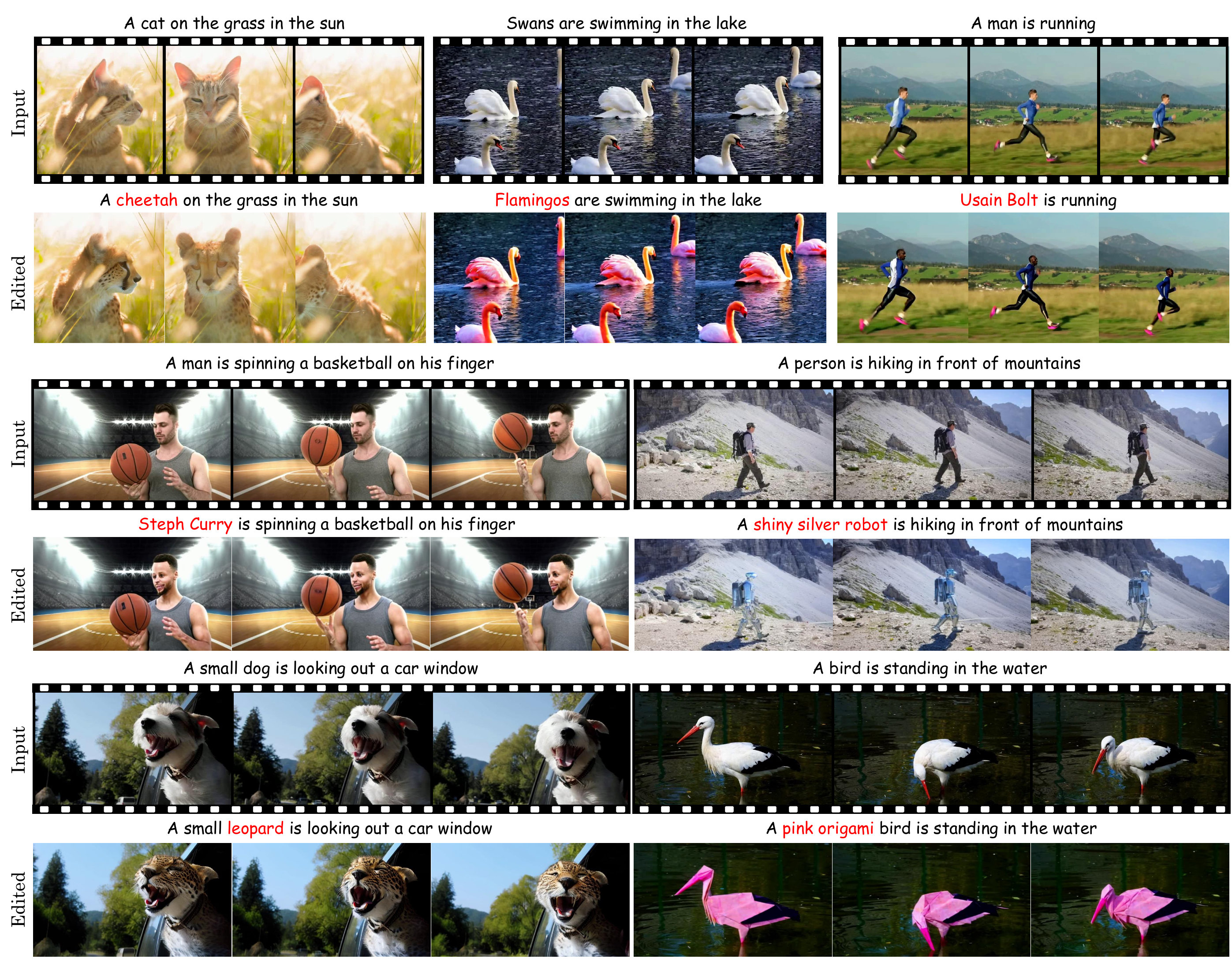}
\caption{\textbf{Slicedit Results.} Our method edits only the specified regions of the input video according to the target prompt while keeping the unspecified regions the same. The output video maintains coherence between frames (\textit{i.e.}, the same robot, origami rabbit and cheetah across the frames). Video results are available on our \href{https://matankleiner.github.io/slicedit/\#examples}{website.}}
\label{fig:results}
\end{figure*}

\paragraph{Spatiotemporal slices.} 
We observe that Stable Diffusion can produce images that share similarities with spatiotemporal slices of real videos. This observation is illustrated in Fig.~\ref{fig:temporal_slices}, where $\mathbf{y}-t$ planes from randomly chosen natural videos are presented alongside images generated using Stable Diffusion. This suggests that spatiotemporal slice images exist within the generative manifold of Stable Diffusion so that the same pretrained T2I denoiser, $\epsilon_\theta(\cdot,p)$, can effectively function as a denoiser for spatiotemporal slices. 
This observation is supported by Fig.~\ref{fig:MSE_sd_denoiser}, which shows that the MSE achieved by the pretrained Stable Diffusion denoiser when applied to spatiotemporal slices (yellow) is comparable to (and even lower than) the MSE it achieves for video frames (blue). For comparison, when applying the denoiser on out-of-distribution data like randomly permuted frames, the MSE is much higher (green). 
For more details see App.~\ref{sm:denoiser_exp}.
With this observation, we apply the denoiser $\epsilon_\theta(\cdot,p)$ separately on each $\mathbf{y}-t$ slice, utilizing an empty prompt $p$ for conditioning (making the denoiser unconditional). This choice aligns with the observations demonstrated in Fig.~\ref{fig:temporal_slices} and is supported by an ablation study provided in App.~\ref{sm:ablation}. We refer to this spatiotemporal denoiser as $\epsilon^{\text{S}}_\theta(\cdot,p)$.
Note that the original denoiser $\epsilon_\theta(\cdot,p)$ operates over representations of dimension $(64,64,4)$, implying that $\epsilon^{\text{S}}_\theta(\cdot,p)$ can only be applied to videos with 64 frames. We explain how to handle shorter and longer videos in Sec.~\ref{sec:experiments}. 

\paragraph{Combined zero-shot video denoiser.} Intuitively, the extended attention denoiser, $\epsilon^{\text{EA}}_\theta(\cdot,p)$ that operates on frames produces a spatially coherent volume, while the spatiotemporal denoiser, $\epsilon^\text{S}_\theta(\cdot,`` \: ")$, generates a temporally consistent volume. Thus, a combination of both should enforce both spatial coherence and temporal consistency. 
Following this intuition, we set the combined video denoiser to be
\begin{equation}
\label{eq:epsilon_volume}
\epsilon^\text{V}_\theta(\cdot,p) = \sqrt{\gamma}\epsilon^{\text{EA}}_\theta(\cdot,p) +  \sqrt{1-\gamma}\epsilon^\text{S}_\theta(\cdot,`` \: ").
\end{equation}
The hyperparameter $\gamma$ balances between the two denoisers while preserving the variance of the predicted noise. Note that each denoiser serves a distinct purpose, hence requiring different prompts. The right pane of Fig.~\ref{fig:overview} depicts this combined denoiser. 

\subsection{Video Editing}

To edit an input video, $I_0$, we adopt the recently proposed DDPM inversion method of \citet{hubermanspiegelglas23}. We first invert the entire video volume by extracting the noise volumes for all diffusion timesteps using $\epsilon^\text{V}_\theta(\cdot,p_{\text{src}})$. Note that this step is fundamentally different from methods that use per-frame inversion \cite{geyer2024tokenflow}. Then, we perform DDPM sampling with $\epsilon^\text{V}_\theta(\cdot,p_{\text{tar}})$, while using the noise volumes obtained from the inversion. 
Note that $p_{\text{src}}$ and $p_{\text{tar}}$ are prompts that describe the input video and the desired edited video, respectively. Therefore, they are fed in $\epsilon^{\text{EA}}_\theta(\cdot,p)$. As noted earlier, the empty prompt in $\epsilon^{\text{S}}_\theta(\cdot,`` \: ")$ is used in the inversion as well as in the sampling processes. 
In addition, during the sampling process we inject the extended attention maps of the source video into the extended attention maps of the edited video. This injection, inspired by~\citet{Tumanyan2023PnP}, helps to preserve the structure and motion of the original video in the edited result. While reminiscent of the approach employed in FateZero~\citep{QI2023FateZero}, our methodology differs in the treatment of attention maps. Unlike the blending process in FateZero, our method involves directly copying attention maps from the input video into the edited one. Figure~\ref{fig:overview} depicts an overview of the entire video editing process. Our editing algorithm is summarized in App.~\ref{sm:algorithmVideo}.

\section{Experiments}
\label{sec:experiments}
\subsection{Implementation Details}

 \begin{figure*}[t]
\centering
\includegraphics[width=\textwidth]{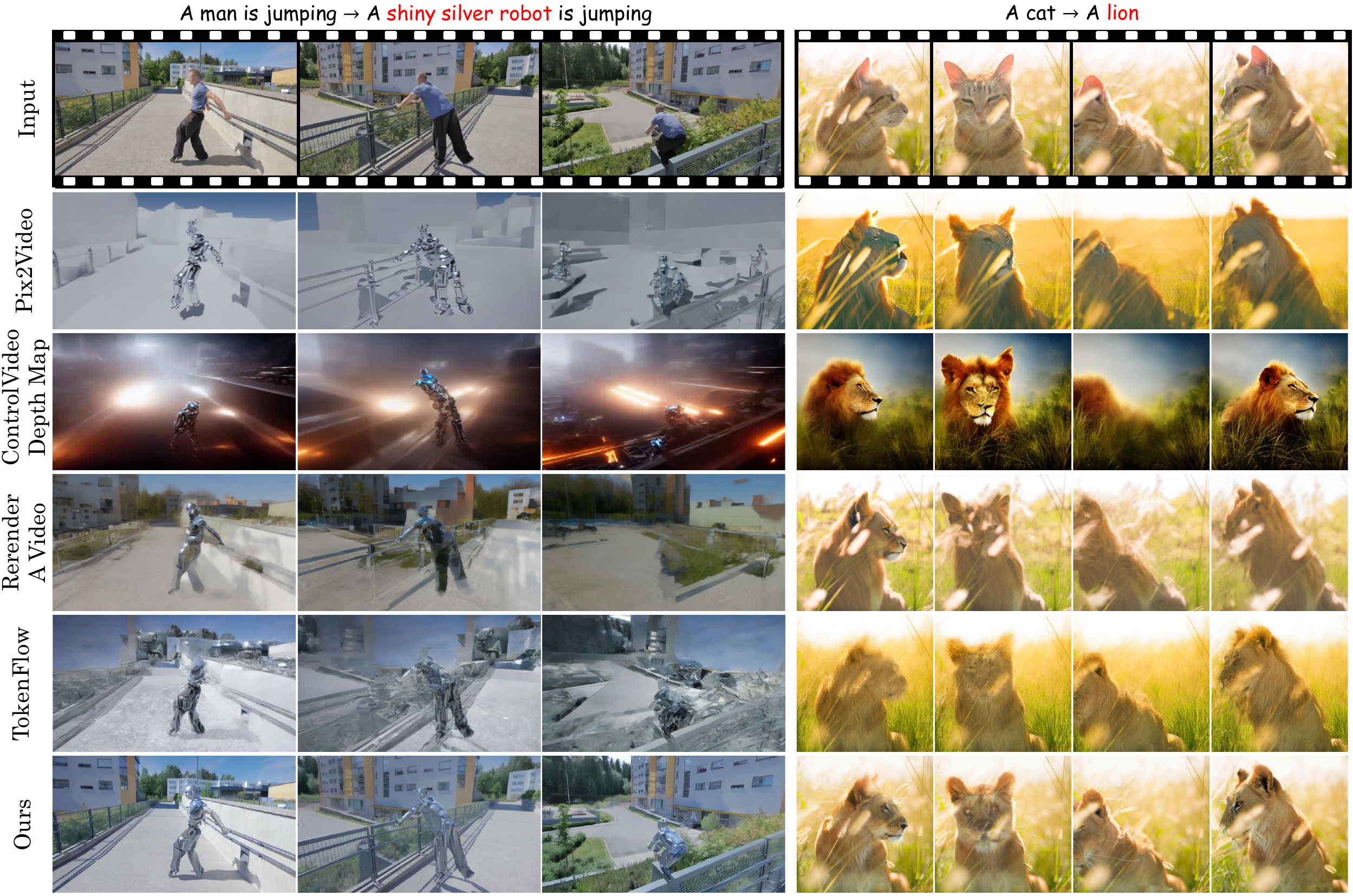}
\caption{\textbf{Qualitative comparison.} We compare our method against other state-of-the-art zero-shot video editing methods. Our method edits only the specified region, according to the text prompt, and keeps the unspecified regions unchanged. The competing methods often change the entire frame, specified and unspecified regions alike. See videos on our \href{https://matankleiner.github.io/slicedit/\#comparisons}{website.}} 
\label{fig:comparison}
\end{figure*}

We use the official weights of Stable Diffusion v2.1\footnote{\url{https://huggingface.co/stabilityai/stable-diffusion-2-1}} as our pretrained T2I denoiser. As mentioned in Sec.~\ref{sec:method}, this model's latent space dimensions are $(64, 64, 4)$. Thus, naive application of our method is possible only for videos with 64 frames. For videos with less than 64 frames (and more than 32), we use RIFE~\citep{huang2022rife}, a state-of-the-art frame interpolation method, to double the number of frames. After editing, we temporally subsample the video back to the original number of frames. For videos exceeding 64 frames, we split the video volume into overlapping segments. The extent of this overlap is determined by the total number of frames. At every diffusion step, we average, while preserving variance, the predicted noise volumes from the overlapping segments. 

We use DDPM inversion with $T=50$ steps. For the generation process, we use DDPM sampling starting from timestep $T-T_{\text{skip}}$, where in all our experiments we fix $T_{\text{skip}}=8$. The parameter $T_{\text{skip}}$ controls the extent to which the edited video adheres to the input video. We set the classifier free guidance~\citep{ho2021classifier} strength parameter to $10$ in $\epsilon^{\text{EA}}$ and to $1$ in $\epsilon^{\text{S}}$. Moreover, we inject the extended attention features from the source video to the target video in 85\% of the sampling process.
We set $\gamma$, the balancing parameter in Eq.~(\ref{eq:epsilon_volume}), to $0.8$. All results in the paper, in the SM, and on our \href{https://matankleiner.github.io/slicedit/\#examples}{website} were produced with these hyperparameters.  
For an ablation study on our design
choices please refer to Sec.~\ref{sec:ablation_study} and App.~\ref{sm:ablation}. The effect of different hyperparameter configurations is discussed in App.~\ref{sm:hyperparams}.

\subsection{Comparisons}
\paragraph{Dataset.} We evaluate our method on a dataset of videos, which we collected from the DAVIS dataset~\citep{pont2017davis}, the LOVEU-TGVE dataset~\citep{wu2023loveucvpr} and from the internet. These videos vary in length and aspect ratio and they depict animals, objects and humans, exhibiting different types of motion.   
Their lengths vary between 32 and 210 frames. Our dataset comprises 60 text-video pairs, for which we manually specified the source and target prompts. We use this dataset to evaluate all competing methods. Some methods remove a few frames from the end of the input video. For fair comparison we ensure all edited videos have the same number of frames, by removing frames from the results of other methods.

 \begin{figure*}[t]
\centering
\includegraphics[width=\linewidth]{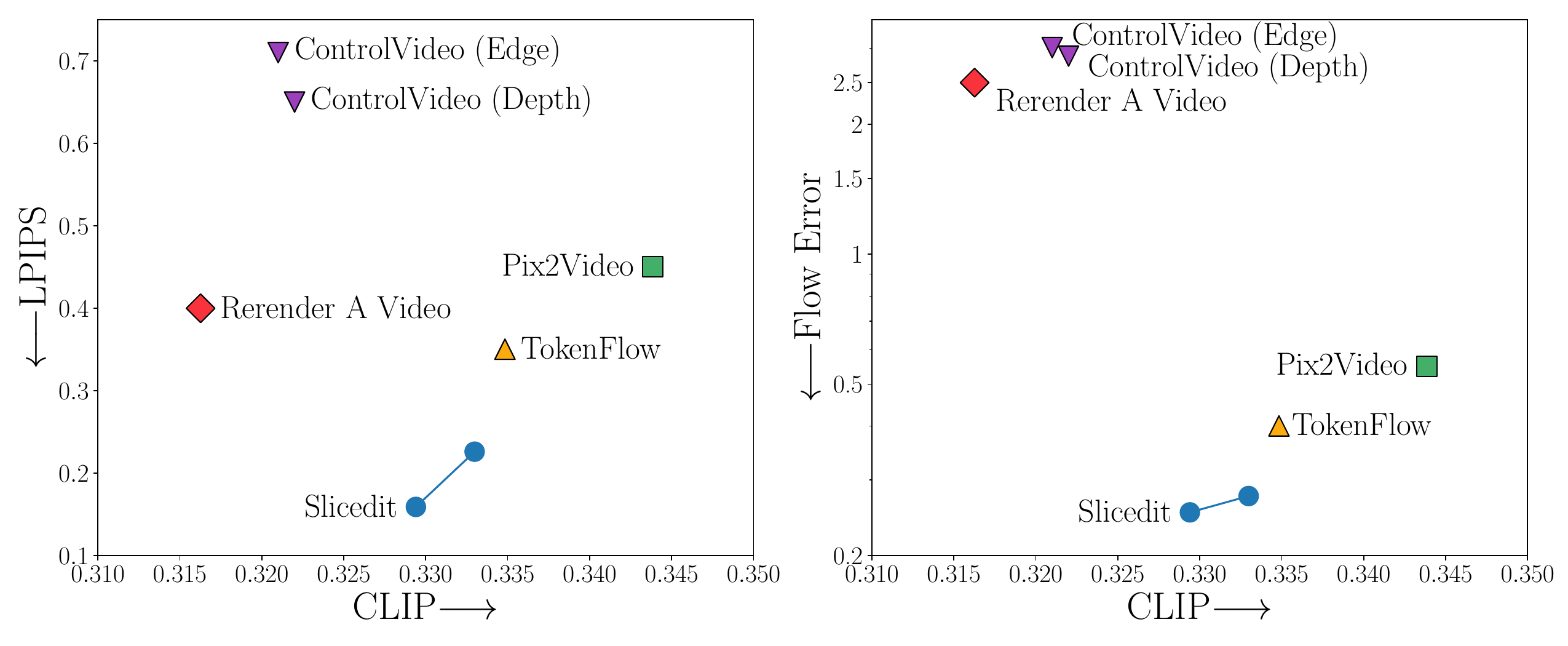}
\caption{\textbf{Numerical comparison to competing methods.} We compare our method and to the competing methods on 60 text-video pairs. The left pane shows the editing fidelity, measured via the CLIP score, vs.~the faithfulness to the original video, measured with LPIPS. The right pane shows the editing fidelity vs.~temporal consistency, measured via flow error. Slicedit is shown with two options for the hyperparameters $T_{\text{skip}}$, classifier-free guidance strength, and percent of extended-attention injection. From left to right, the values of these parameters are (8,10,85) and (8,14,85). The first setting is the one used to produce all the results in the paper, appendix, and the website.}
\label{fig:metrics}
\end{figure*}

\paragraph{Competing methods.} We compare our method against state-of-the-art recent and concurrent methods, whose code is publicly available at the time of writing. 
The competing methods are Pix2Video~\cite{Ceylan2023Pix2Video}, Rerender A Video~\citep{Yang23Rerender}, TokenFlow~\citep{geyer2024tokenflow} and ControlVideo~\citep{zhang2024controlvideo}. ControlVideo aims to generate a new video based on a text prompt and a conditioning signal extracted from the original video, \textit{e.g.} depth maps and edge maps. Therefore, their method generates videos that adhere only to the motion and structure of the original video, but do not preserve its textures and colors. More details about the competing methods can be found in App.~\ref{sm:comparisons}. 
We excluded Text2Video Zero~\citep{Khachatryan2023text2video-zero} and FateZero~\citep{QI2023FateZero} from our comparisons due to their memory requirements. On a single RTX A6000 GPU, the one we used for running all methods including ours, these methods could edit videos of only up to 30 frames. Additionally, Tune A Video~\citep{Wu23TuneAVideo} and Flatten~\citep{cong2024flatten} were omitted due to their extended processing time and memory requirements, respectively, which do not permit large-scale comparisons. A qualitative comparison to both these methods can be found in Figs.~\ref{fig:SM_comps},~\ref{fig:SM_comps_2}.  

\subsection{Qualitative Evaluation}
Figures~\ref{fig:teaser} and~\ref{fig:results} present editing results obtained with Slicedit on challenging videos involving camera motion and complex nonrigid object motion, including occlusions. In all cases, our method manages to successfully edit the input videos according to the text prompt. 

Figure~\ref{fig:comparison} presents a comparison between our method and the competing methods. For ControlVideo, we used depth conditioning. Our method adheres to the text prompt while doing a better job than the other methods at preserving the unspecified regions. This is true for both complex motion (left), and for smooth motion (right). 

Specifically, in the case of the parkour video, Pix2Video, Rerender A Video and ControlVideo all exhibit major inconsistencies in both the edited robot and the background. In the result of Rerender A Video, the robot fades away in the last part of the edited video, as can be seen in the right frame. TokenFlow's edit is somewhat more successful, yet the robot is jittery and blurry. In addition, it can be clearly seen that all competing methods drastically change the background, even though the target prompt only refers to the person. Pix2Video and TokenFlow turn the background into a gray, CGI-styled city, a change not specified in the text prompt. Rerender A Video, while retaining some of the background's color palette, also makes significant changes.

In the case of the cat video, which exhibits smoother motion, Rerender A Video and TokenFlow generate blurry images while Pix2Vid and ControlVideo show inconsistency in the motion. More frame comparisons are provided in App.~\ref{sm:visual-comaprisons} and video comparisons appear on our \href{https://matankleiner.github.io/slicedit/\#comparisons}{website.} 

\subsection{Quantitative Evaluation}
\label{sec:quan_eval}
We conduct a numerical evaluation using metrics that quantify editing fidelity, structure preservation, and temporal consistency. 
Following prior works, we evaluate the editing fidelity using the average cosine similarity between the image CLIP embeddings of the frames and the text CLIP embedding of the target prompt. For structure preservation, we report the LPIPS distance between the source and edited frames. 
The very high LPIPS score of ControlVideo is due to its conditioning only on a signal (\textit{e.g.} depth or edges) from an existing video. To assess temporal consistency, we compute the error between the optical flow fields of the source and edited videos. Specifically, we compute optical flow using RAFT~\cite{teed2020raft} between every pair of consecutive frames in both the original and edited videos. Subsequently, we calculate mean $L_2$ distance between these flow fields, while removing pixels with left-right inconsistencies in the source video. Importantly, unlike the popular warping error~\citep{lai2018learning, Ceylan2023Pix2Video, geyer2024tokenflow}, which is computed by warping the edited frames using the flow field of the original one, the flow error we use captures only motion preservation, and is completely disentangled from appearance inconsistencies.

\begin{figure}[t]
\centering
\includegraphics[width=\linewidth]{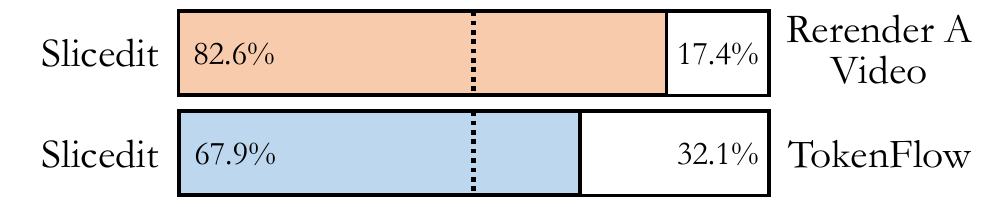}
\caption{\textbf{User study.} We report the percentage of users who preferred our method over Rerender A Video and TokenFlow, when answering which edited video best preserves the essence of the original video.}
\label{fig:user_study}
\end{figure}

Figure~\ref{fig:metrics} presents our quantitative results compared to the competing methods. We show results for our method with two parameter configurations, which lead to different balances between text adherence and fidelity to the input video in motion and appearance (see Tab.~\ref{tab:SM_hyperparams}). Our zero-shot editing demonstrates superior LPIPS scores and flow error, successfully adhering to the text prompt for video editing. In contrast, the competing methods exhibit video edits with some loss of fidelity to the original content and poorer temporal consistency.

We additionally evaluated our method via a user study, in which each participant was shown the original video, the target prompt, and our edited result next to a competing edited result. The order between the two edited results was random. Users were instructed to select the edited video that best preserves the essence of the original video. A screenshot from the user study can be found in Fig.~\ref{fig:sm_user_study}. 
We conducted two separate user studies through Amazon Mechanical Turk, where 50 workers compared our method against TokenFlow and Rerender A Video over our entire dataset. The user study result, reported in Fig.~\ref{fig:user_study}, reveals that humans clearly prefer our edited videos over those of the competing methods in terms of preservation of unspecified regions.

\subsection{Ablation Study}
\label{sec:ablation_study}

 \begin{figure}[t]
\centering
\includegraphics[width=\linewidth]{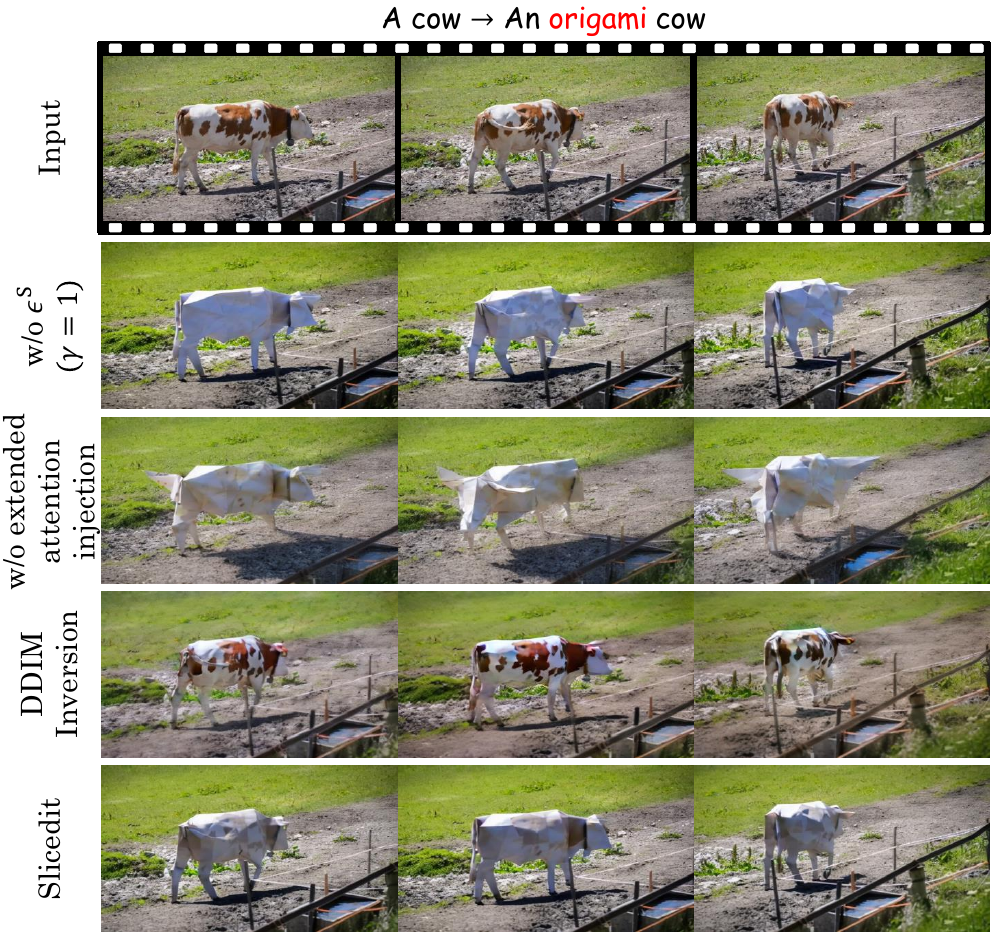}
\caption{\textbf{Ablation study.} Each row displays the results without a key component of our method. The last row, displays the results achieved by Slicedit. Video comparisons are available on our \href{https://matankleiner.github.io/slicedit/\#ablation}{website.}}
\label{fig:ablation}
\end{figure}

 \begin{figure}
\centering
\includegraphics[width=\linewidth]{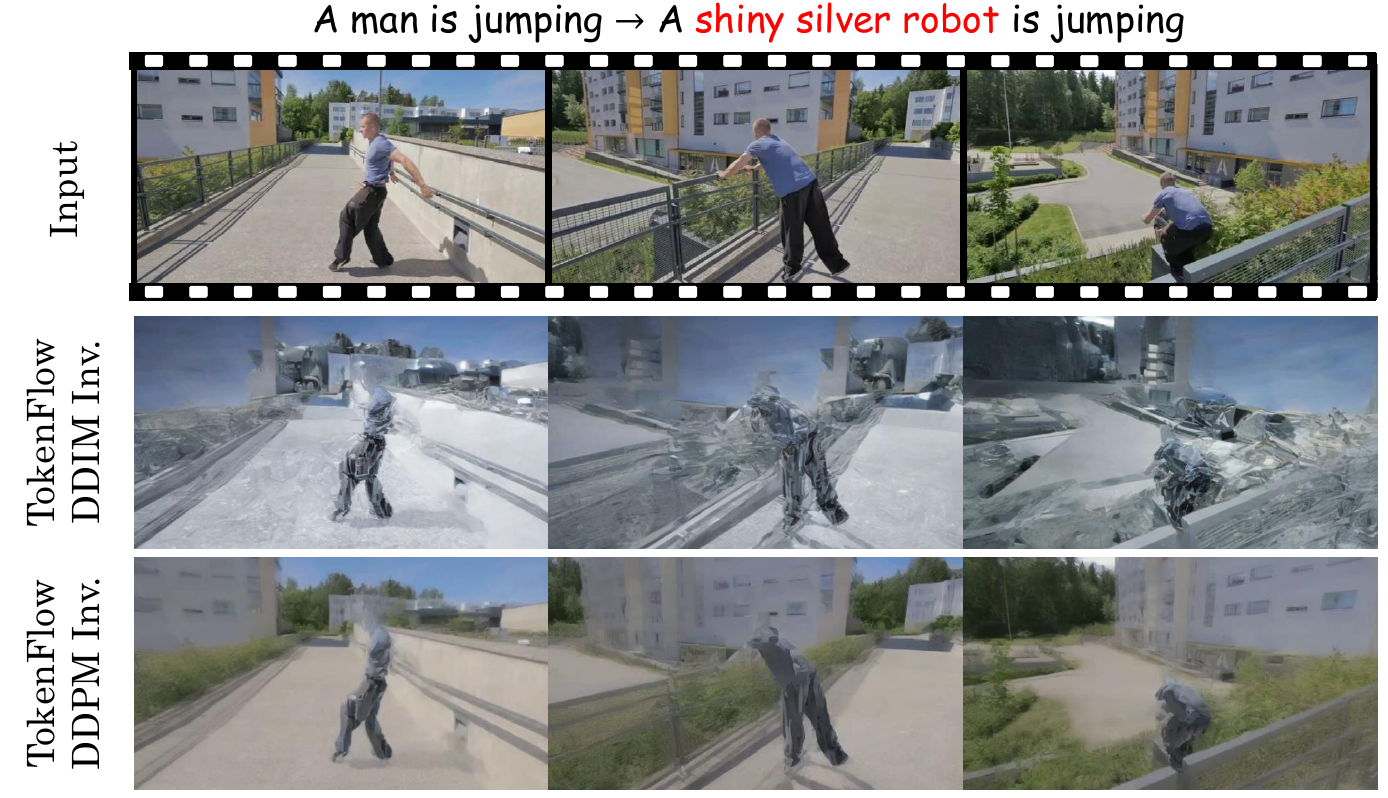}
\caption{\textbf{Volume inversion ablation.} TokenFlow results for DDPM and DDIM per-frame inversion. Videos are available on our \href{https://matankleiner.github.io/slicedit/\#ablation}{website.}}
\label{fig:ablation_volume}
\end{figure}

We next study the importance of each of the components that we used for improving temporal consistency and preserving structure: (1)~denoising spatio-temporal slices, (2)~extended attention injection, (3)~DDPM inversion over DDIM inversion, and (4)~inversion for the entire volume rather than per-frame. 
We note that the incorporation of extended attention in the inflated denoiser also contributes to the final results, however this mechanism is already well established and has been ablated in previous works. As such, we will not provide an ablation for it here. 

As can be seen in Fig.~\ref{fig:ablation}, removing the spatiotemporal denoising (second row) harms the temporal consistency of the edited video, causing the edited object to change appearance over frames. Without injecting the extended attention maps of the original video  (third row), the resulting edited video is not loyal to the layout and motion of the original video. When using DDIM instead of DDPM inversion with the same parameters (fourth row) our method is not able to successfully edit the video according to the text prompt.

For evaluating the effect of performing volume rather than per-frame inversion, it is not feasible to use our inflated denoiser. Hence, we study this with the TokenFlow method, which inherently relies on per-frame inversion. Figure~\ref{fig:ablation_volume} shows the results of TokenFlow with per-frame DDIM and with per-frame DDPM inversion. As can be seen, both cases lead to blurry results. This highlights the importance of performing volume inversion.

For a more detailed ablation, including changing the text prompt of the spatio-temporal denoiser and inversion with an empty source text, comparison of quantitative metrics for each configuration and more frame comparisons, see App.~\ref{sm:ablation}. Video comparisons are available on our \href{https://matankleiner.github.io/slicedit/\#ablation}{website.}

\section{Conclusion}

 \begin{figure}
\centering
\includegraphics[width=0.9\linewidth]{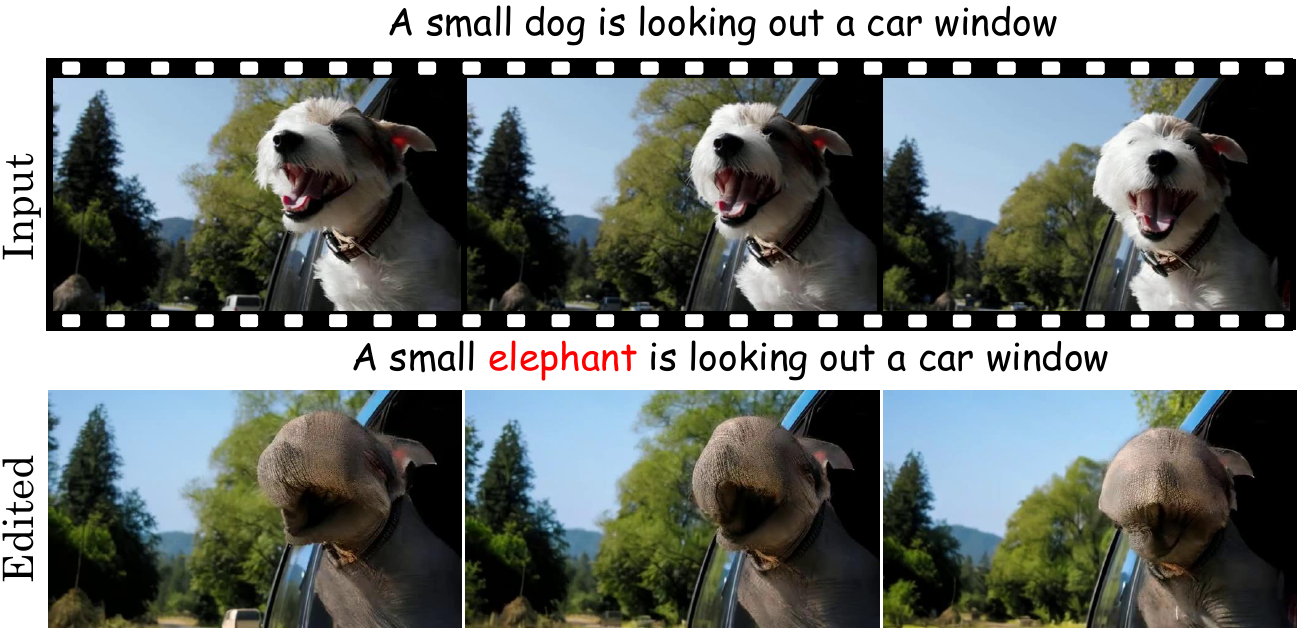}
\caption{\textbf{Failure cases.}  Our method is limited to structure preserving edits, and cannot modify \eg a dog into an elephant. Failure case videos are available on our \href{https://matankleiner.github.io/slicedit/\#ablation}{website.}}
\label{fig:limitation}
\end{figure}

We introduced Slicedit, a zero-shot text-based video editing method utilizing a pretrained text-to-image diffusion model. Our method inflates the model to work on videos using several modifications. Most importantly, it applies the pretrained denoiser, initially designed for images, also on spatiotemporal slices of the video.
To edit videos, we use our inflated denoiser in a DDPM inversion process, in conjunction with injection of the extended attention from the source video to the target video. Our method outperforms existing techniques, successfully editing the video while preserving the unspecified regions without compromising on temporal consistency. We evaluated it by measuring editing fidelity, structure reservation, and temporal consistency metrics, supplemented by a user study. While our method excels at preserving the structure of the input video, it encounters challenges with global editing tasks, such as converting the frames of a natural video into paintings. In addition, our method is limited to structure-preserving edits. This stems from using DDPM inversion with attention injection. An example failure case is shown in Fig.~\ref{fig:limitation}.

\section*{Societal Impact}

Our goal in this work is to suggest a different method for leveraging T2I foundation models for video editing, overcoming some of the weaknesses of previous and concurrent works. Our and similar methods may potentially be used to edit videos to create fake or harmful content. We believe that it is crucial to develop and apply tools for detecting videos edited using generative AI methods, such as ours. In addition, mitigating biases and NSFW content from large datasets, used for training foundation models, will also contribute for safer generative AI usage, including using T2I models for video editing.      

\paragraph{Acknowledgements.} This research was supported by the Israel Science Foundation (grant no. 2318/22) and by the Ollendorff Minerva Center, ECE faculty, Technion. Nathaniel Cohen's stay at the Technion was supported in part by the Mines Paris Foundation.

\bibliography{0_main}

\begin{thebibliography}{43}
\providecommand{\natexlab}[1]{#1}
\providecommand{\url}[1]{\texttt{#1}}
\expandafter\ifx\csname urlstyle\endcsname\relax
  \providecommand{\doi}[1]{doi: #1}\else
  \providecommand{\doi}{doi: \begingroup \urlstyle{rm}\Url}\fi

\bibitem[Adelson \& Bergen(1985)Adelson and Bergen]{adelson1985spatiotemporal}
Adelson, E.~H. and Bergen, J.~R.
\newblock Spatiotemporal energy models for the perception of motion.
\newblock \emph{Josa a}, 2\penalty0 (2):\penalty0 284--299, 1985.

\bibitem[Betker et~al.(2023)Betker, Goh, Jing, Brooks, Wang, Li, Ouyang, Zhuang, Lee, Guo, et~al.]{betker2023improving}
Betker, J., Goh, G., Jing, L., Brooks, T., Wang, J., Li, L., Ouyang, L., Zhuang, J., Lee, J., Guo, Y., et~al.
\newblock Improving image generation with better captions.
\newblock \emph{Computer Science. https://cdn. openai. com/papers/dall-e-3. pdf}, 2023.

\bibitem[Bolles et~al.(1987)Bolles, Baker, and Marimont]{bolles1987epipolar}
Bolles, R.~C., Baker, H.~H., and Marimont, D.~H.
\newblock Epipolar-plane image analysis: An approach to determining structure from motion.
\newblock \emph{International journal of computer vision}, 1\penalty0 (1):\penalty0 7--55, 1987.

\bibitem[Brooks et~al.(2023)Brooks, Holynski, and Efros]{Brooks22InstructPix2Pix}
Brooks, T., Holynski, A., and Efros, A.~A.
\newblock Instruct{P}ix{2}pix: Learning to follow image editing instructions.
\newblock In \emph{Proceedings of the {IEEE} Conference on Computer Vision and Pattern Recognition (CVPR)}, pp.\  18392--18402, 2023.

\bibitem[Canny(1986)]{canny1986computational}
Canny, J.
\newblock A computational approach to edge detection.
\newblock \emph{IEEE Transactions on pattern analysis and machine intelligence}, \penalty0 (6):\penalty0 679--698, 1986.

\bibitem[Ceylan et~al.(2023)Ceylan, Huang, and Mitra]{Ceylan2023Pix2Video}
Ceylan, D., Huang, C.-H.~P., and Mitra, N.~J.
\newblock Pix2video: Video editing using image diffusion.
\newblock In \emph{Proceedings of the IEEE/CVF International Conference on Computer Vision (ICCV)}, pp.\  23206--23217, October 2023.

\bibitem[Cong et~al.(2024)Cong, Xu, christian simon, Chen, Ren, Xie, Perez-Rua, Rosenhahn, Xiang, and He]{cong2024flatten}
Cong, Y., Xu, M., christian simon, Chen, S., Ren, J., Xie, Y., Perez-Rua, J.-M., Rosenhahn, B., Xiang, T., and He, S.
\newblock {FLATTEN}: optical {FL}ow-guided {ATTEN}tion for consistent text-to-video editing.
\newblock In \emph{The Twelfth International Conference on Learning Representations}, 2024.
\newblock URL \url{https://openreview.net/forum?id=JgqftqZQZ7}.

\bibitem[Dai et~al.(2023)Dai, Hou, Ma, Tsai, Wang, Wang, Zhang, Vandenhende, Wang, Dubey, et~al.]{dai2023emu}
Dai, X., Hou, J., Ma, C.-Y., Tsai, S., Wang, J., Wang, R., Zhang, P., Vandenhende, S., Wang, X., Dubey, A., et~al.
\newblock Emu: Enhancing image generation models using photogenic needles in a haystack.
\newblock \emph{arXiv preprint arXiv:2309.15807}, 2023.

\bibitem[Esser et~al.(2023)Esser, Chiu, Atighehchian, Granskog, and Germanidis]{esser2023structure}
Esser, P., Chiu, J., Atighehchian, P., Granskog, J., and Germanidis, A.
\newblock Structure and content-guided video synthesis with diffusion models.
\newblock In \emph{Proceedings of the IEEE/CVF International Conference on Computer Vision}, pp.\  7346--7356, 2023.

\bibitem[Feng et~al.(2023)Feng, Weng, Wang, Yuan, Bao, Luo, Chen, and Guo]{feng2023ccedit}
Feng, R., Weng, W., Wang, Y., Yuan, Y., Bao, J., Luo, C., Chen, Z., and Guo, B.
\newblock Ccedit: Creative and controllable video editing via diffusion models.
\newblock \emph{arXiv preprint arXiv:2309.16496}, 2023.

\bibitem[Geyer et~al.(2024)Geyer, Bar-Tal, Bagon, and Dekel]{geyer2024tokenflow}
Geyer, M., Bar-Tal, O., Bagon, S., and Dekel, T.
\newblock Tokenflow: Consistent diffusion features for consistent video editing.
\newblock In \emph{The Twelfth International Conference on Learning Representations}, 2024.
\newblock URL \url{https://openreview.net/forum?id=lKK50q2MtV}.

\bibitem[Hertz et~al.(2023)Hertz, Mokady, Tenenbaum, Aberman, Pritch, and Cohen{-}Or]{Hertz22p2p}
Hertz, A., Mokady, R., Tenenbaum, J., Aberman, K., Pritch, Y., and Cohen{-}Or, D.
\newblock Prompt-to-prompt image editing with cross-attention control.
\newblock In \emph{International Conference on Learning Representations (ICLR)}, 2023.

\bibitem[Ho \& Salimans(2021)Ho and Salimans]{ho2021classifier}
Ho, J. and Salimans, T.
\newblock Classifier-free diffusion guidance.
\newblock In \emph{NeurIPS 2021 Workshop on Deep Generative Models and Downstream Applications}, 2021.

\bibitem[Ho et~al.(2020)Ho, Jain, and Abbeel]{ho2020denoising}
Ho, J., Jain, A., and Abbeel, P.
\newblock Denoising diffusion probabilistic models, 2020.

\bibitem[Huang et~al.(2022)Huang, Zhang, Heng, Shi, and Zhou]{huang2022rife}
Huang, Z., Zhang, T., Heng, W., Shi, B., and Zhou, S.
\newblock Real-time intermediate flow estimation for video frame interpolation.
\newblock In \emph{European Conference on Computer Vision}, pp.\  624--642. Springer, 2022.

\bibitem[Huberman-Spiegelglas et~al.(2023)Huberman-Spiegelglas, Kulikov, and Michaeli]{hubermanspiegelglas23}
Huberman-Spiegelglas, I., Kulikov, V., and Michaeli, T.
\newblock An edit friendly {DDPM} noise space: Inversion and manipulations, 2023.

\bibitem[Kawar et~al.(2023)Kawar, Zada, Lang, Tov, Chang, Dekel, Mosseri, and Irani]{Bahjat22Imagic}
Kawar, B., Zada, S., Lang, O., Tov, O., Chang, H., Dekel, T., Mosseri, I., and Irani, M.
\newblock Imagic: Text-based real image editing with diffusion models.
\newblock In \emph{Conference on Computer Vision and Pattern Recognition (CVPR)}, 2023.

\bibitem[Khachatryan et~al.(2023)Khachatryan, Movsisyan, Tadevosyan, Henschel, Wang, Navasardyan, and Shi]{Khachatryan2023text2video-zero}
Khachatryan, L., Movsisyan, A., Tadevosyan, V., Henschel, R., Wang, Z., Navasardyan, S., and Shi, H.
\newblock Text2video-zero: Text-to-image diffusion models are zero-shot video generators.
\newblock In \emph{Proceedings of the IEEE/CVF International Conference on Computer Vision (ICCV)}, pp.\  15954--15964, October 2023.

\bibitem[Lai et~al.(2018)Lai, Huang, Wang, Shechtman, Yumer, and Yang]{lai2018learning}
Lai, W.-S., Huang, J.-B., Wang, O., Shechtman, E., Yumer, E., and Yang, M.-H.
\newblock Learning blind video temporal consistency.
\newblock In \emph{Proceedings of the European conference on computer vision (ECCV)}, pp.\  170--185, 2018.

\bibitem[Lin et~al.(2017)Lin, Feng, dos Santos, Yu, Xiang, Zhou, and Bengio]{lin2017selfattention}
Lin, Z., Feng, M., dos Santos, C.~N., Yu, M., Xiang, B., Zhou, B., and Bengio, Y.
\newblock A {STRUCTURED} {SELF}-{ATTENTIVE} {SENTENCE} {EMBEDDING}.
\newblock In \emph{International Conference on Learning Representations}, 2017.
\newblock URL \url{https://openreview.net/forum?id=BJC_jUqxe}.

\bibitem[Liu et~al.(2023)Liu, Zhang, Li, Lin, and Jia]{liu2023video-p2p}
Liu, S., Zhang, Y., Li, W., Lin, Z., and Jia, J.
\newblock Video-p2p: Video editing with cross-attention control.
\newblock \emph{arXiv preprint arXiv:2303.04761}, 2023.

\bibitem[Ngo et~al.(2003)Ngo, Pong, and Zhang]{ngo2003motion}
Ngo, C.-W., Pong, T.-C., and Zhang, H.-J.
\newblock Motion analysis and segmentation through spatio-temporal slices processing.
\newblock \emph{IEEE Transactions on Image Processing}, 12\penalty0 (3):\penalty0 341--355, 2003.

\bibitem[Pont-Tuset et~al.(2017)Pont-Tuset, Perazzi, Caelles, Arbel{\'a}ez, Sorkine-Hornung, and Van~Gool]{pont2017davis}
Pont-Tuset, J., Perazzi, F., Caelles, S., Arbel{\'a}ez, P., Sorkine-Hornung, A., and Van~Gool, L.
\newblock The 2017 davis challenge on video object segmentation.
\newblock \emph{arXiv preprint arXiv:1704.00675}, 2017.

\bibitem[QI et~al.(2023)QI, Cun, Zhang, Lei, Wang, Shan, and Chen]{QI2023FateZero}
QI, C., Cun, X., Zhang, Y., Lei, C., Wang, X., Shan, Y., and Chen, Q.
\newblock Fatezero: Fusing attentions for zero-shot text-based video editing.
\newblock In \emph{Proceedings of the IEEE/CVF International Conference on Computer Vision (ICCV)}, pp.\  15932--15942, October 2023.

\bibitem[Radford et~al.(2021)Radford, Kim, Hallacy, Ramesh, Goh, Agarwal, Sastry, Askell, Mishkin, Clark, et~al.]{radford2021learning}
Radford, A., Kim, J.~W., Hallacy, C., Ramesh, A., Goh, G., Agarwal, S., Sastry, G., Askell, A., Mishkin, P., Clark, J., et~al.
\newblock Learning transferable visual models from natural language supervision.
\newblock In \emph{International conference on machine learning}, pp.\  8748--8763. PMLR, 2021.

\bibitem[Ranftl et~al.(2021)Ranftl, Bochkovskiy, and Koltun]{ranftl2021vision}
Ranftl, R., Bochkovskiy, A., and Koltun, V.
\newblock Vision transformers for dense prediction.
\newblock In \emph{Proceedings of the IEEE/CVF international conference on computer vision}, pp.\  12179--12188, 2021.

\bibitem[Rav-Acha et~al.(2005)Rav-Acha, Pritch, Lischinski, and Peleg]{rav2005dynamosaics}
Rav-Acha, A., Pritch, Y., Lischinski, D., and Peleg, S.
\newblock Dynamosaics: Video mosaics with non-chronological time.
\newblock In \emph{2005 IEEE Computer Society Conference on Computer Vision and Pattern Recognition (CVPR'05)}, volume~1, pp.\  58--65. IEEE, 2005.

\bibitem[Rombach et~al.(2022)Rombach, Blattmann, Lorenz, Esser, and Ommer]{Rombach22LDM}
Rombach, R., Blattmann, A., Lorenz, D., Esser, P., and Ommer, B.
\newblock High-resolution image synthesis with latent diffusion models.
\newblock In \emph{Proceedings of the {IEEE} Conference on Computer Vision and Pattern Recognition (CVPR)}, pp.\  10684--10695, 2022.

\bibitem[Ronneberger et~al.(2015)Ronneberger, Fischer, and Brox]{ronneberger2015unet}
Ronneberger, O., Fischer, P., and Brox, T.
\newblock U-net: Convolutional networks for biomedical image segmentation, 2015.

\bibitem[Saharia et~al.(2022)Saharia, Chan, Saxena, Li, Whang, Denton, Ghasemipour, Gontijo~Lopes, Karagol~Ayan, Salimans, et~al.]{saharia2022photorealistic}
Saharia, C., Chan, W., Saxena, S., Li, L., Whang, J., Denton, E.~L., Ghasemipour, K., Gontijo~Lopes, R., Karagol~Ayan, B., Salimans, T., et~al.
\newblock Photorealistic text-to-image diffusion models with deep language understanding.
\newblock \emph{Advances in Neural Information Processing Systems}, 35:\penalty0 36479--36494, 2022.

\bibitem[Song et~al.(2020)Song, Meng, and Ermon]{song2020denoising}
Song, J., Meng, C., and Ermon, S.
\newblock Denoising diffusion implicit models.
\newblock In \emph{International Conference on Learning Representations}, 2020.

\bibitem[Teed \& Deng(2020)Teed and Deng]{teed2020raft}
Teed, Z. and Deng, J.
\newblock Raft: Recurrent all-pairs field transforms for optical flow.
\newblock In \emph{Computer Vision--ECCV 2020: 16th European Conference, Glasgow, UK, August 23--28, 2020, Proceedings, Part II 16}, pp.\  402--419. Springer, 2020.

\bibitem[Tumanyan et~al.(2023)Tumanyan, Geyer, Bagon, and Dekel]{Tumanyan2023PnP}
Tumanyan, N., Geyer, M., Bagon, S., and Dekel, T.
\newblock Plug-and-play diffusion features for text-driven image-to-image translation.
\newblock In \emph{Proceedings of the IEEE/CVF Conference on Computer Vision and Pattern Recognition (CVPR)}, pp.\  1921--1930, June 2023.

\bibitem[Vaswani et~al.(2017)Vaswani, Shazeer, Parmar, Uszkoreit, Jones, Gomez, Kaiser, and Polosukhin]{vaswani2017attention}
Vaswani, A., Shazeer, N., Parmar, N., Uszkoreit, J., Jones, L., Gomez, A.~N., Kaiser, {\L}., and Polosukhin, I.
\newblock Attention is all you need.
\newblock \emph{Advances in neural information processing systems}, 30, 2017.

\bibitem[Vu \& Chandler(2014)Vu and Chandler]{vu2014vis}
Vu, P.~V. and Chandler, D.~M.
\newblock Vis 3: an algorithm for video quality assessment via analysis of spatial and spatiotemporal slices.
\newblock \emph{Journal of Electronic Imaging}, 23\penalty0 (1):\penalty0 013016--013016, 2014.

\bibitem[Wang et~al.(2023)Wang, Xie, Liu, Chen, Cao, Wang, and Shen]{Wang2023vid2vid-zero}
Wang, W., Xie, k., Liu, Z., Chen, H., Cao, Y., Wang, X., and Shen, C.
\newblock Zero-shot video editing using off-the-shelf image diffusion models.
\newblock \emph{arXiv preprint arXiv:2303.17599}, 2023.

\bibitem[Wu et~al.(2023{\natexlab{a}})Wu, Ge, Wang, Lei, Gu, Shi, Hsu, Shan, Qie, and Shou]{Wu23TuneAVideo}
Wu, J.~Z., Ge, Y., Wang, X., Lei, S.~W., Gu, Y., Shi, Y., Hsu, W., Shan, Y., Qie, X., and Shou, M.~Z.
\newblock Tune-a-video: One-shot tuning of image diffusion models for text-to-video generation.
\newblock In \emph{Proceedings of the {IEEE} International Conference on Computer Vision (ICCV)}, pp.\  7623--7633, October 2023{\natexlab{a}}.

\bibitem[Wu et~al.(2023{\natexlab{b}})Wu, Li, Gao, Dong, Bai, Singh, Xiang, Li, Huang, Sun, et~al.]{wu2023loveucvpr}
Wu, J.~Z., Li, X., Gao, D., Dong, Z., Bai, J., Singh, A., Xiang, X., Li, Y., Huang, Z., Sun, Y., et~al.
\newblock Cvpr 2023 text guided video editing competition.
\newblock \emph{arXiv preprint arXiv:2310.16003}, 2023{\natexlab{b}}.

\bibitem[Xing et~al.(2023)Xing, Dai, Hu, Wu, and Jiang]{xing2023simda}
Xing, Z., Dai, Q., Hu, H., Wu, Z., and Jiang, Y.-G.
\newblock Simda: Simple diffusion adapter for efficient video generation.
\newblock \emph{arXiv preprint arXiv:2308.09710}, 2023.

\bibitem[Yang et~al.(2023)Yang, Zhou, Liu, , and Loy]{Yang23Rerender}
Yang, S., Zhou, Y., Liu, Z., , and Loy, C.~C.
\newblock Rerender a video: Zero-shot text-guided video-to-video translation.
\newblock In \emph{ACM SIGGRAPH Asia Conference Proceedings}, 2023.

\bibitem[Zhang et~al.(2023)Zhang, Rao, and Agrawala]{zhang2023adding}
Zhang, L., Rao, A., and Agrawala, M.
\newblock Adding conditional control to text-to-image diffusion models.
\newblock In \emph{Proceedings of the IEEE/CVF International Conference on Computer Vision}, pp.\  3836--3847, 2023.

\bibitem[Zhang et~al.(2024)Zhang, Wei, Jiang, ZHANG, Zuo, and Tian]{zhang2024controlvideo}
Zhang, Y., Wei, Y., Jiang, D., ZHANG, X., Zuo, W., and Tian, Q.
\newblock Controlvideo: Training-free controllable text-to-video generation.
\newblock In \emph{The Twelfth International Conference on Learning Representations}, 2024.
\newblock URL \url{https://openreview.net/forum?id=5a79AqFr0c}.

\bibitem[Zuckerman et~al.(2020)Zuckerman, Naor, Pisha, Bagon, and Irani]{zuckerman2020across}
Zuckerman, L.~P., Naor, E., Pisha, G., Bagon, S., and Irani, M.
\newblock Across scales and across dimensions: Temporal super-resolution using deep internal learning.
\newblock In \emph{Computer Vision--ECCV 2020: 16th European Conference, Glasgow, UK, August 23--28, 2020, Proceedings, Part VII 16}, pp.\  52--68. Springer, 2020.

\end{thebibliography}
\bibliographystyle{icml2024}

\clearpage

\onecolumn

\renewcommand\thefigure{S\arabic{figure}}    
\setcounter{figure}{0}  
\renewcommand{\thesection}{\Alph{section}}
\setcounter{section}{0}
\renewcommand{\thetable}{S\arabic{table}}
\setcounter{table}{0}
\renewcommand{\theequation}{S\arabic{equation}}
\setcounter{equation}{0}

\section{Additional Results}

\begin{figure}[h]
\centering
\includegraphics[width=0.98\textwidth]{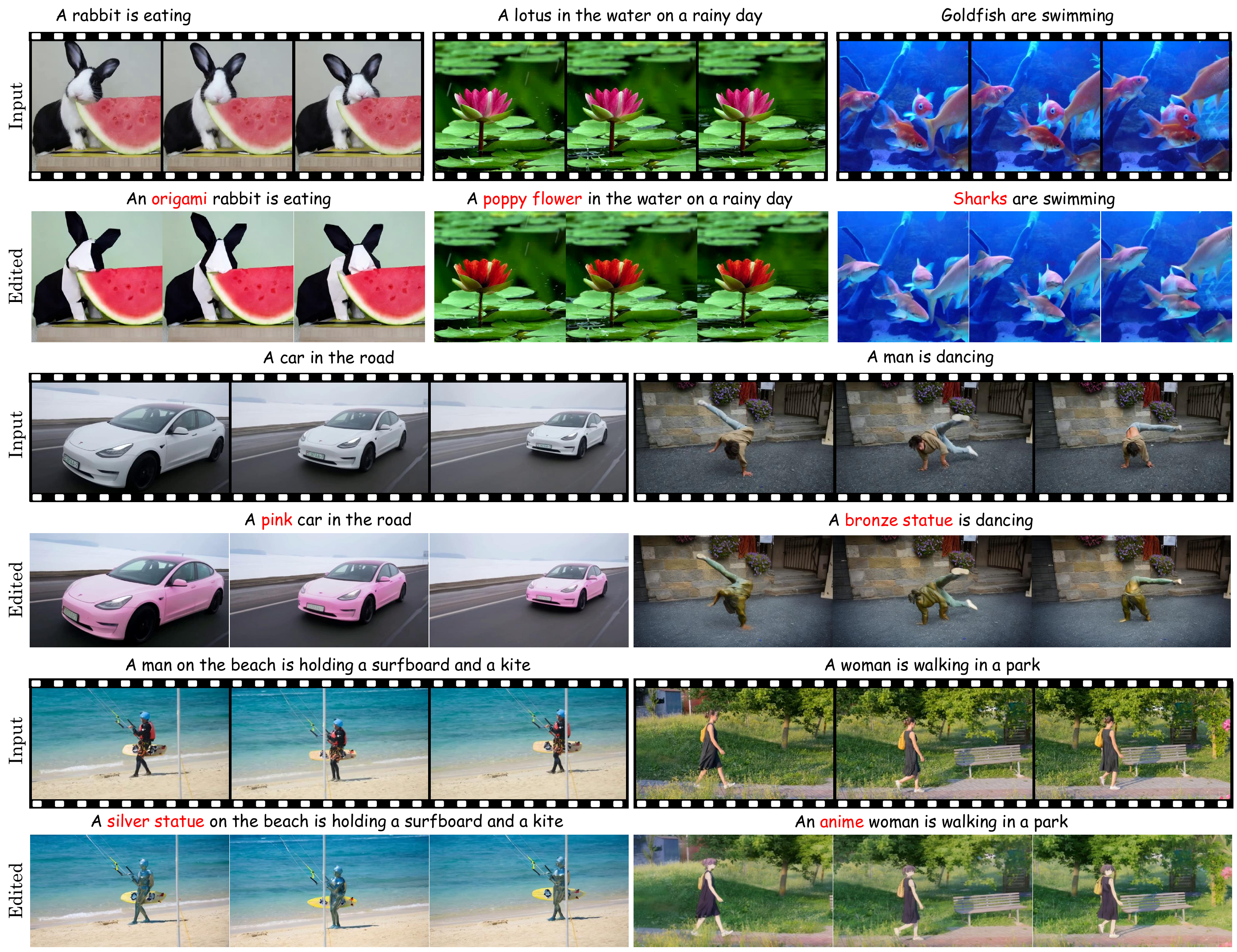}
\caption{\textbf{Additional editing results using our method.}}
\label{fig:SM_results}
\end{figure}

Fig.~\ref{fig:SM_results} displays additional results for our video editing. 

\clearpage

\section{Additional Visual Comparisons}
\label{sm:visual-comaprisons}
\begin{figure}[h]
\centering
\includegraphics[width=0.95\textwidth]{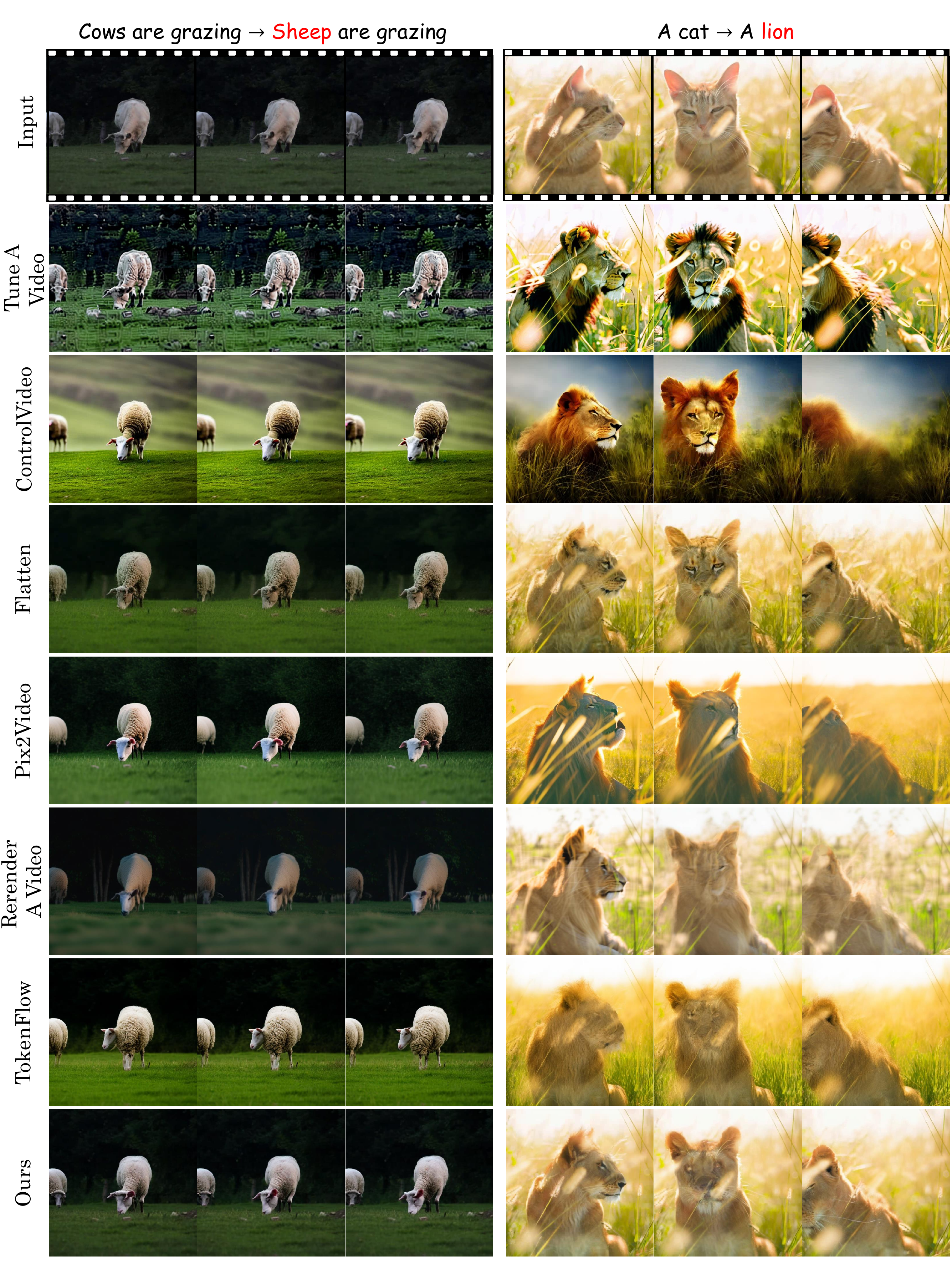}
\caption{\textbf{Additional qualitative comparisons.}}
\label{fig:SM_comps}
\end{figure}

\begin{figure}[h]
\centering
\includegraphics[width=0.95\textwidth]{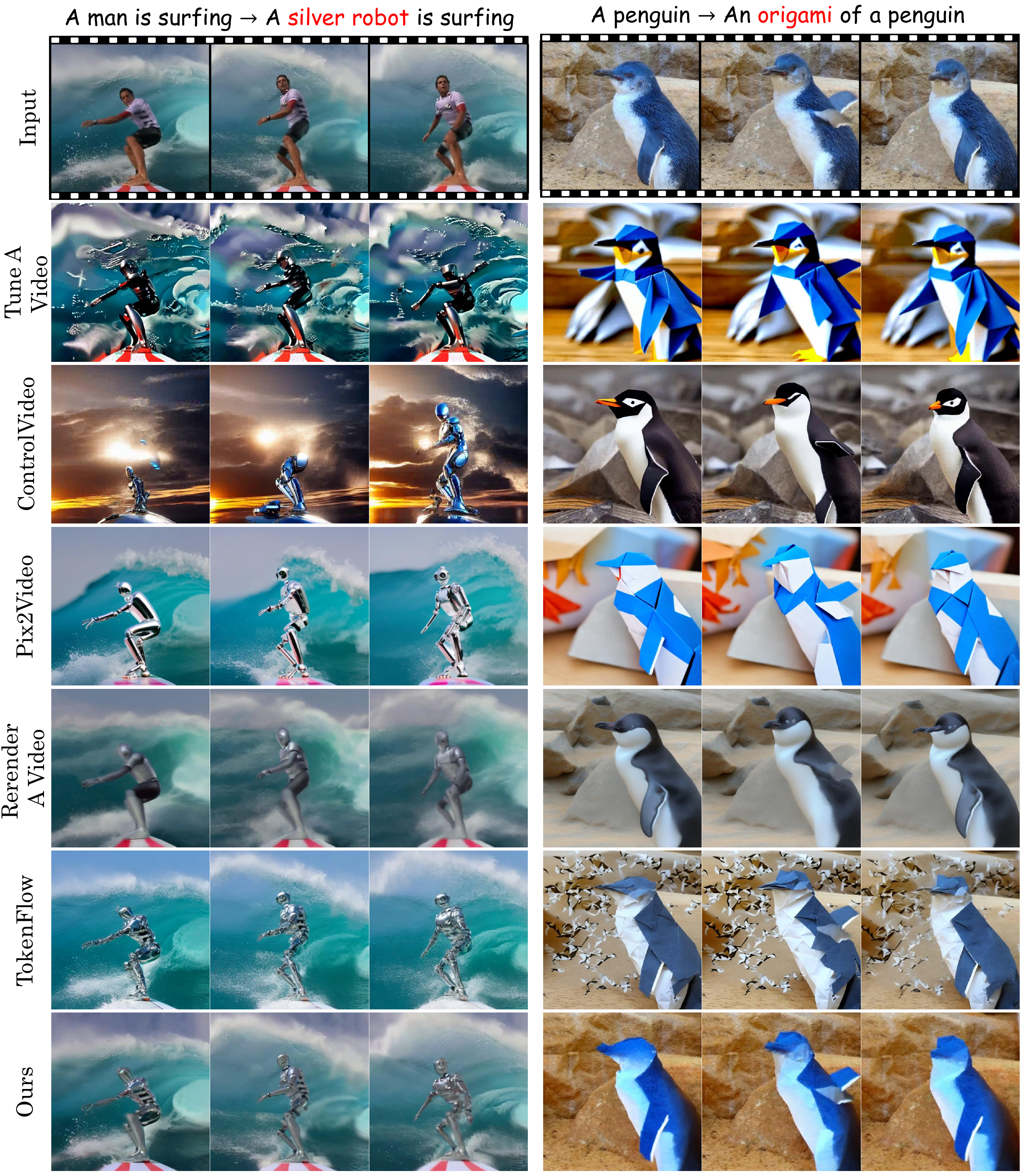}
\caption{\textbf{Additional qualitative comparisons.}}
\label{fig:SM_comps_2}
\end{figure}

As mentioned in Sec.~\ref{sec:experiments}, we omitted Tune A Video from the quantitative comparison due to their extended processing time, as shown in Tab.~\ref{tab:time}. Here we provide a qualitative comparison for this method on 4 different text-video pairs. As can be seen in Figs.~\ref{fig:SM_comps},~\ref{fig:SM_comps_2}, Tune A Video is able to adhere to a text prompt through optimization. However, this optimization process often results in undesired global changes in the edited video. For example, the photorealistic waves surrounding the surfing man in the right pane of Fig.~\ref{fig:SM_comps} turn into painted-like waves.

We also provide here quantitative comparisons to Flatten for videos with 32 frames, which can be seen in Fig.~\ref{fig:SM_comps}. Flatten results keep the unspecified regions similar to the original video. For the cat video, Flatten edit did not keep the identity of the lion across different frames. The lion's face in the middle frame is very similar to the original cat's, and is quite different from the lion's face on the right frame. As Flatten's memory requirement made it not possible to edit videos longer than 32 frames on a single RTX A6000 GPU, which we used for our method and all competing methods, we do not provide here more results of this method.

ControlVideo aims to generate a video based on a text prompt and a condition signal from the original video. Therefore, its results adhere only to the original video motion and structure, but not to the original video colors and and textures. As a result, the unspecified regions in their results are always different than the original video's, as can be seen in all the figures below,
Here we provide qualitative results for ControlVideo with depth maps conditioning. For depth estimation we used MiDaS DPT-Hybrid~\citep{ranftl2021vision}, as in the ControlVideo paper. Quantitative results for canny edge~\citep{canny1986computational} conditioning are available at Fig.~\ref{fig:metrics} and at Tab.~\ref{tab:SM_quantitative}. 

\begin{figure}[h]
\centering
\includegraphics[width=\textwidth]{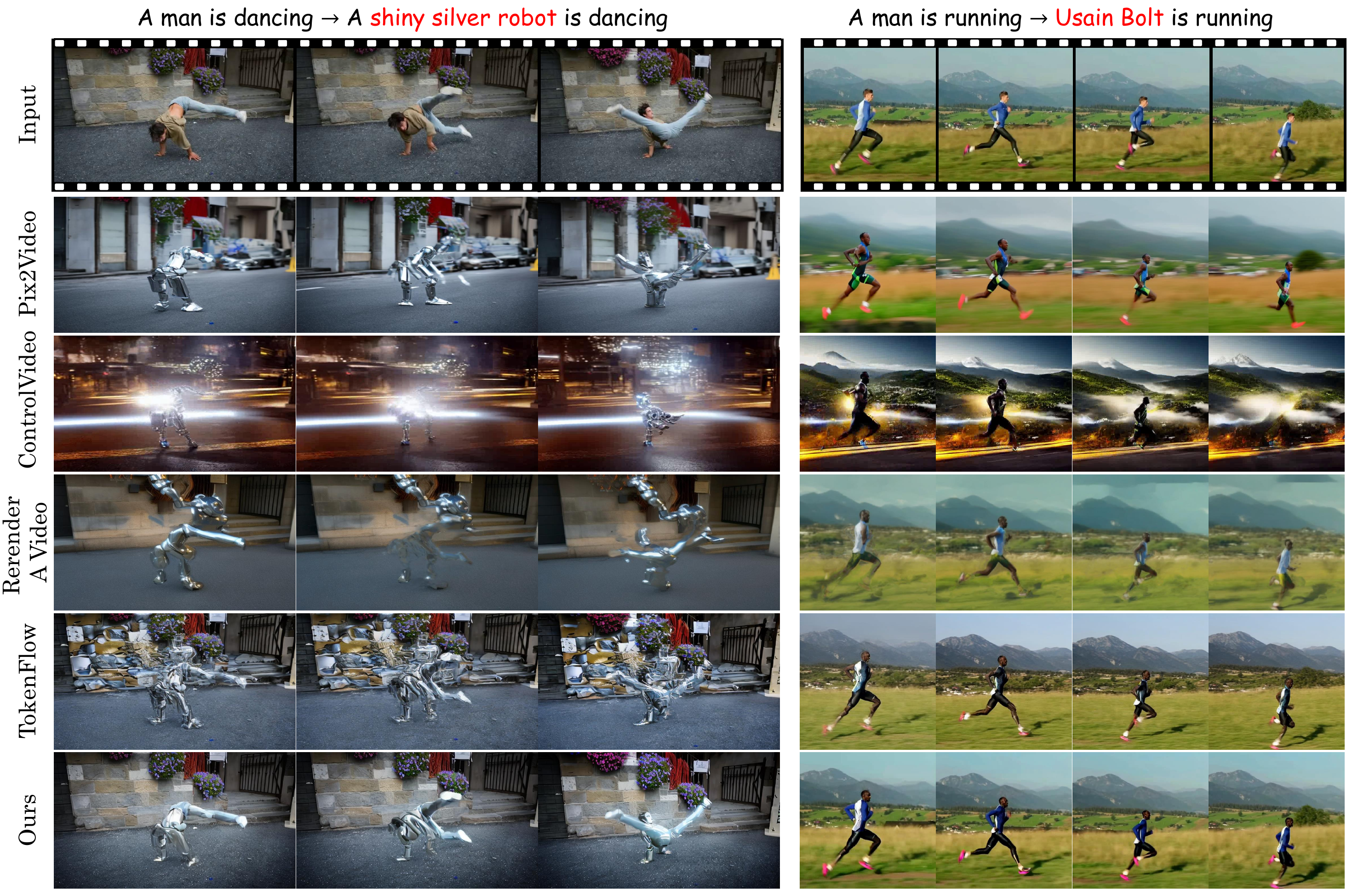}
\caption{\textbf{Additional qualitative comparisons.}}
\label{fig:SM_comps_1}
\end{figure}

\begin{figure}[h]
\centering
\includegraphics[width=\textwidth]{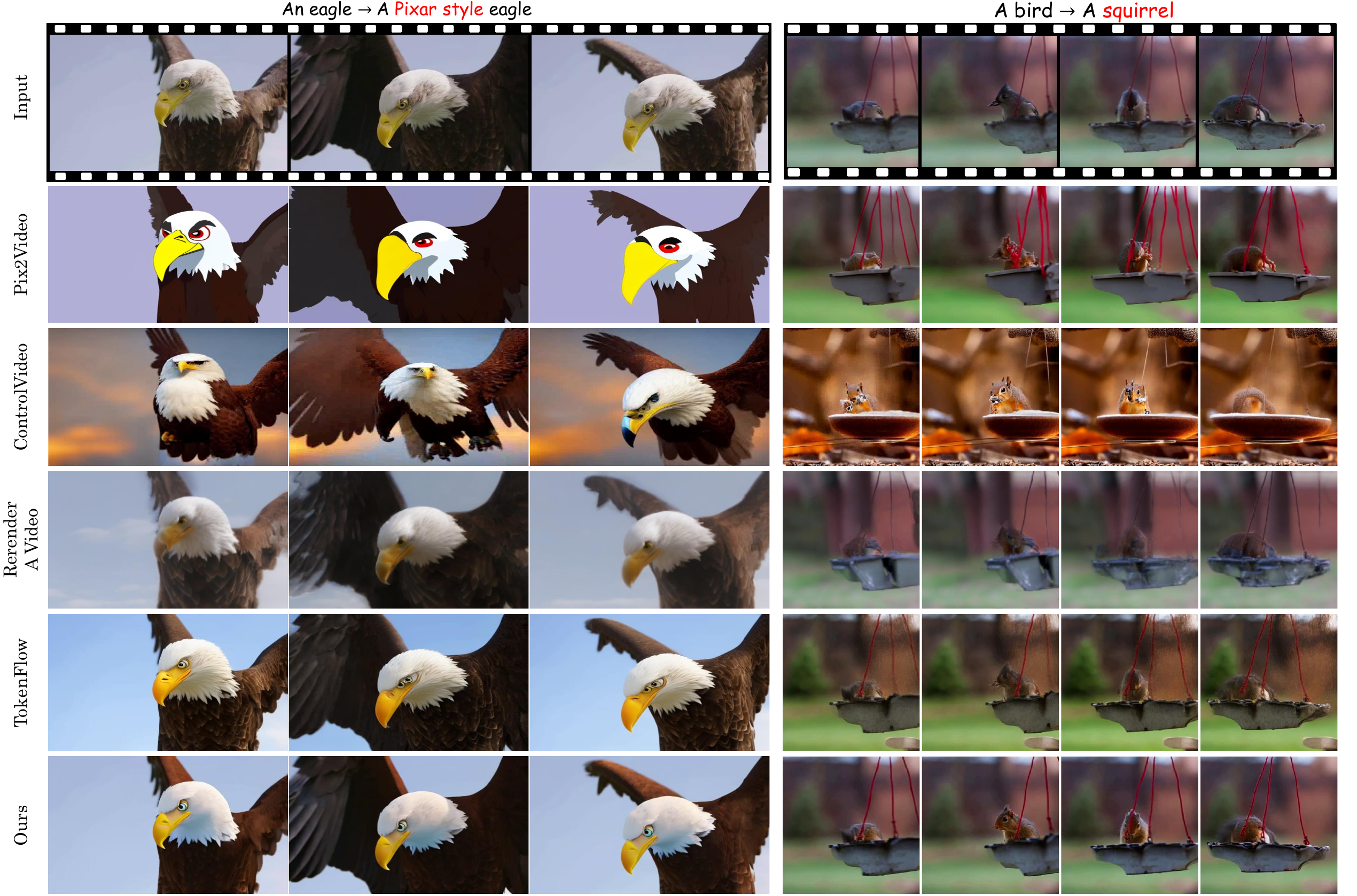}
\caption{\textbf{Additional qualitative comparisons.}}
\label{fig:SM_eagle_comps}
\end{figure}

Figs.~\ref{fig:SM_comps_1}-\ref{fig:SM_comps_2} also illustrate that the competing methods often introduce global changes to the entire video, resulting in editing of regions unrelated to the target text together with the related regions. These global changes almost always include color changes, illustrated by the eagle example in Fig.~\ref{fig:SM_eagle_comps} and the cows example in Fig.~\ref{fig:SM_comps}. It also includes changing objects' shapes as the grass in the cat example in Fig.~\ref{fig:SM_comps}, and changing the background. Background changes can be seen in the dancing man example in Fig.~\ref{fig:SM_comps_1} and the penguin example in Fig.~\ref{fig:SM_comps_2}, where the competing methods completely changed the background. More subtle changes to the background also occurs, as in the running man example in Fig.~\ref{fig:SM_comps_1}, where all competing methods changed the village behind the running man.

It can also be noticed that some of the competing methods failed to adhere to text prompt (Pix2Video in the eagle eaxmple in Fig.~\ref{fig:SM_eagle_comps}, TokenFlow in the bird example in Fig.~\ref{fig:SM_eagle_comps}), to the original video motion (TokenFlow in the cows example in Fig.~\ref{fig:SM_comps}), or produce inconsistent editing or blurry output (ControlVideo in Figs.~\ref{fig:SM_comps_1},~\ref{fig:SM_eagle_comps}, Rerender A Video and TokenFlow in the cat example in Fig.~\ref{fig:SM_comps}).   

Video results of the comparisons can be found on our~\href{https://matankleiner.github.io/slicedit/\#comparisons}{website.}

\clearpage
\section{Comparisons}
\label{sm:comparisons}
\subsection{Competing Methods}

We compare our method with competing methods whose official implementation was publicly available at the time of writing: Pix2Video, TokenFlow, Rerender A Video and ControlVideo. 
Pix2Video~\cite{Ceylan2023Pix2Video} changes the self attention of a pretrained T2I model into extended attention. It also uses depth estimation as part of the input to the T2I model. TokenFlow~\citep{geyer2024tokenflow} and Rerender A Video~\citep{Yang23Rerender} edit key frames and propagate their features to the rest of the frames in a zero-shot manner. Rerender A Video also utilizes optical flow and edge maps to guide the editing process. ControlVideo~\citep{zhang2024controlvideo} utilize a trained ControlNet~\cite{zhang2023adding} for controllable text to video generation, using a text prompt and condition signal from the original video. We compared here to ControlVideo with depth maps and canny edge~\citep{canny1986computational} conditioning, as suggested in their paper. We used MiDaS DPT-Hybrid~\citep{ranftl2021vision} for depth estimation.

We followed the instructions provided by the authors in their official implementation for setting and executing their code.
ControlVideo For Pix2Video, ControlVideo and TokenFlow we used the hyperparameters described in their paper and used in their official implementation. For TokenFlow it included sampling with negative prompt. For ControlVideo it included sampling with positive prompt and negative prompt. 

As detailed in their paper, Rerender A Video tunes their editing hyperparameter per video. In addition, their official implementation includes sampling with ``negative'' and ``additive'' prompts that are tuned per video. Additive prompts are prompts that include words that are known to improve T2I model results, as ``high-quality''. As it is impractical in our case to tune hyperparameters for each text-video pair, or negative and additive prompts for each text-video pair, we used the most common hyperparameters found in the official implementation of Rerender A Video, and did not sample with negative or additive prompts.

\subsection{Running Time}

Table~\ref{tab:time} compares the running time of our method against the competing methods. We used the same 64 frames video for all methods, which we executed on the same RTX A6000 GPU. We consider running time as processing time (inversion, tuning) and editing time together. 

The running time of Rerender A Video changes according to the number of key frames. This hyperparameter can be tuned per video. According to their paper, the recommended number of key frames is between 5 to 20, where they used 10 key frames. We followed their selection and used 10 key frames in all our evaluations, as well as in this running time comparison. 

The running time of Flatten is given for 32 frames video, as its memory requirements made it inapplicable to longer videos on the RTX A6000 GPU we used. We marked it by with * in the table.

\begin{table}[h]
    \centering
\caption{\textbf{Running time comparison}}
\label{tab:time}
    \begin{tabular}{|c|ccccccc|} \hline 
         --- &  Tune A Video & ControlVideo & Flatten\textsuperscript{*} & Pix2Video & Rerender A Video & TokenFlow & Ours \\ \hline 
         Time [min] & 359 & 6 & 10 & 15.5 & 9 & 16.25 & 33.7 \\ \hline 
    \end{tabular}
\end{table}

\clearpage

\subsection{Metric Comparison}

In addition to the metrics presented in Sec.~\ref{sec:experiments}, we also measure video coherence using CLIP\footnote{\url{https://github.com/openai/CLIP} with ViT-B/32}~\citep{radford2021learning}. We embed each pair of consecutive frames of the edited video into CLIP image space and report the average cosine similarity of all video frame pairs. This metric can also be computed over the original videos, providing a reference for the typical values associated with a video exhibiting natural coherence.

In the context of Flow error, explained in Sec.~\ref{sec:quan_eval}, we remove pixels that do not match the left-right consistencies calculated over the original video. Specifically, We calculate the distance between the flow in each frame and the subsequent frame as well as the reverse direction. Pixels are excluded from the error calculation if the disparity between the flows exceeds 1 pixel.

The following table includes the metric numerical results as used for creating Fig.~\ref{fig:metrics} in addition to the metric mentioned above. It is worth noting that the CLIP-consistency metric yields nearly identical values across all methods, closely resembling the CLIP consistency of the original videos. 

\begin{table}[h]
    \centering
\caption{\textbf{Quantitative Comparison} }
\label{tab:SM_quantitative}
    \begin{tabular}{|l|cccc|} \hline 
         Method &  CLIP-Text $\uparrow$ & CLIP-Consistency & Flow Err. $\downarrow$ & LPIPS $\downarrow$ \\ \hline\hline 
         Original & --- & 0.982 & --- & --- \\ \hline 
         Pix2Video & 0.343 & 0.982 & 0.55 & 0.45 \\
         Rerender A Video& 0.316 & 0.983 & 2.5 & 0.4\\
         TokenFlow& 0.334 & 0.986  & 0.4 &0.35 \\ 
         ControlVideo (edge)& 0.322 & 0.975  & 2.88 &0.65 \\ 
         ControlVideo (depth)& 0.321 & 0.977  & 3.014 &0.71 \\ 
         Ours& 0.329 & 0.982 & 0.252  & 0.159  \\ \hline
    \end{tabular}
\end{table}

\clearpage

\subsection{User Study}
Below we present a screenshot from our user study interface. The user is given the source video along with our edited video and another competing method's edited video (Rerender A Video or TokenFlow). The user is instructed to choose the edited video that maintains the essence of the original video.
\begin{figure}[h]
\centering
\includegraphics[width=0.85\textwidth]{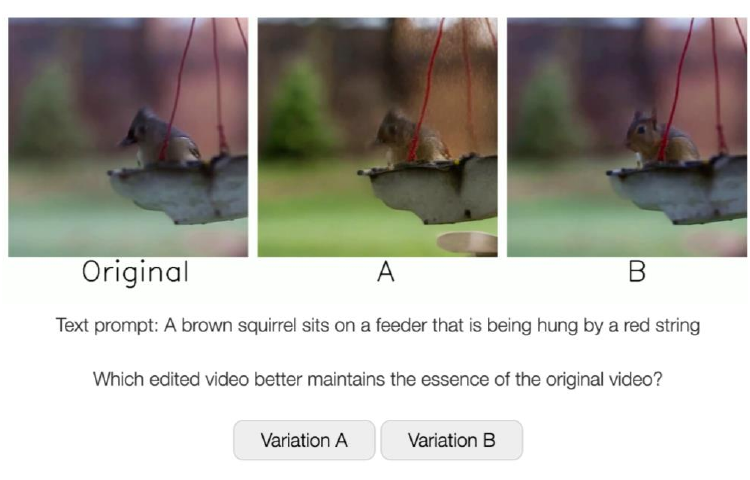}
\caption{\textbf{A screenshot from our user study.}}
\label{fig:sm_user_study}
\end{figure}

\clearpage
\section{Ablation Study}
\label{sm:ablation}
In this section, we analyze the effectiveness of our design choices through an ablation study. Note that whenever we modify a specific aspect of the method, all other components remain unchanged. We evaluate each modification on our dataset (described in Sec.~\ref{sec:experiments}). The quantitative results are summarized in Tab.~\ref{tab:SM_ablation}. 

\subsection{Spatiotemporal Slices}
To justify the utilization of spatiotemporal slices as a component of our zero-shot video editing method, we conducted an ablation study by excluding $\epsilon_{\theta}^\text{S}$ from the inversion and sampling process. Moreover, we extended the ablation study by including a denoiser over $\mathbf{x}-t$ slices in addition to $\mathbf{y}-t$ slices. In this scenario, $\epsilon_{\theta}^\text{S}$ is computed as an average of these two denoisers. We conducted additional experiments by employing a different text denoiser than the empty string. Specifically, the prompts used are depicted in the right pane of Fig.~\ref{fig:temporal_slices}, where $\text{text}_1=$``painting of geological rock folding in sedimentary layers'' and $\text{text}_2=$``motion blur''.

Table~\ref{tab:SM_ablation} shows the effectiveness of the spatiotemporal slices component in our method. Excluding this component results in a higher Flow error (first row). Incorporating the $\mathbf{x}-t$ slices into the calculations does yield a slight improvement in the Flow error and LPIPS score (second row), yet, imposes a longer running time. Finally, the results obtained with different text prompts for $\epsilon_{\theta}^\text{S}$ exhibit minimal influence (third and fourth rows).

\subsection{Inversion and Sampling via DDIM}
We performed an additional ablation by replacing the DDPM inversion with DDIM inversion~\cite{song2020denoising}. The inflated denoiser explained in Sec.~\ref{sec:method} is being used as well as the extended attention injection. However, the inversion and the sampling process use the DDIM schemes. We keep the hyperparameters as before while acknowledging that scanning for the optimal hyperparameters may affect results. The result shown in Tab.~\ref{tab:SM_ablation}, fifth row, highlights the significance of DDPM inversion in our method in editing ability (CLIP-Text score) and temporal consistency (Flow error).

\subsection{Extended Attention Injection}
To assess the effectiveness of injecting extended attention from the source to the target video, we evaluate it by excluding this component from the method. As anticipated, the results in Tab.~\ref{tab:SM_ablation}, sixth row, demonstrate 
the importance of this module in our method for time consistency.

\subsection{Empty Prompt Inversion}
We additionally assess the importance of the source prompt used during the inversion phase. The results, detailed in Table~\ref{tab:SM_ablation} under the seventh row, indicate that they are less optimal.

\begin{table}[h]
\centering
\caption{\textbf{Ablation study.} Evaluation when changing some of the method's design choices. The last row of the table represents the method without any change.}
\label{tab:SM_ablation}
    \begin{tabular}{|l|cccc|} \hline 
          &  CLIP-Text $\uparrow$ & CLIP-Consistency & Flow Err. $\downarrow$ & LPIPS $\downarrow$ \\ \hline\hline 
         Original & --- & 0.982 & --- & --- \\ \hline 
         without $\mathbf{y}-t$ slices ($\gamma=1$) & 0.335 & 0.982 & 0.279 & 0.207  \\
         adding $\mathbf{x}-t$ slice & 0.329 & 0.982 & 0.248 & 0.155  \\
        $\epsilon^\text{S}(\cdot,\text{text}_1)$ & 0.30 & 0.982  & 0.252 & 0.161\\
        $\epsilon^\text{S}(\cdot,\text{text}_2)$ & 0.30 &0.982 & 0.252 & 0.161 \\ 
        \hline 
            DDIM-inversion & 0.29  & 0.97 & 0.55 & 0.19 \\ 
        \hline 
        exclude ext-att injection & 0.348 & 0.982 & 0.452 & 0.292 \\
        \hline 
        empty prompt inversion & 0.323 & 0.983 & 0.253 & 0.182 \\
         \hline
         \hline  
           Ours& 0.329 & 0.982 & 0.252  & 0.159  \\ \hline
     \end{tabular}

\end{table}
\clearpage
\subsection{Visual Illustration of the Effects of our Design Choices}

\begin{figure}[h]
\centering
\includegraphics[width=0.95\textwidth]{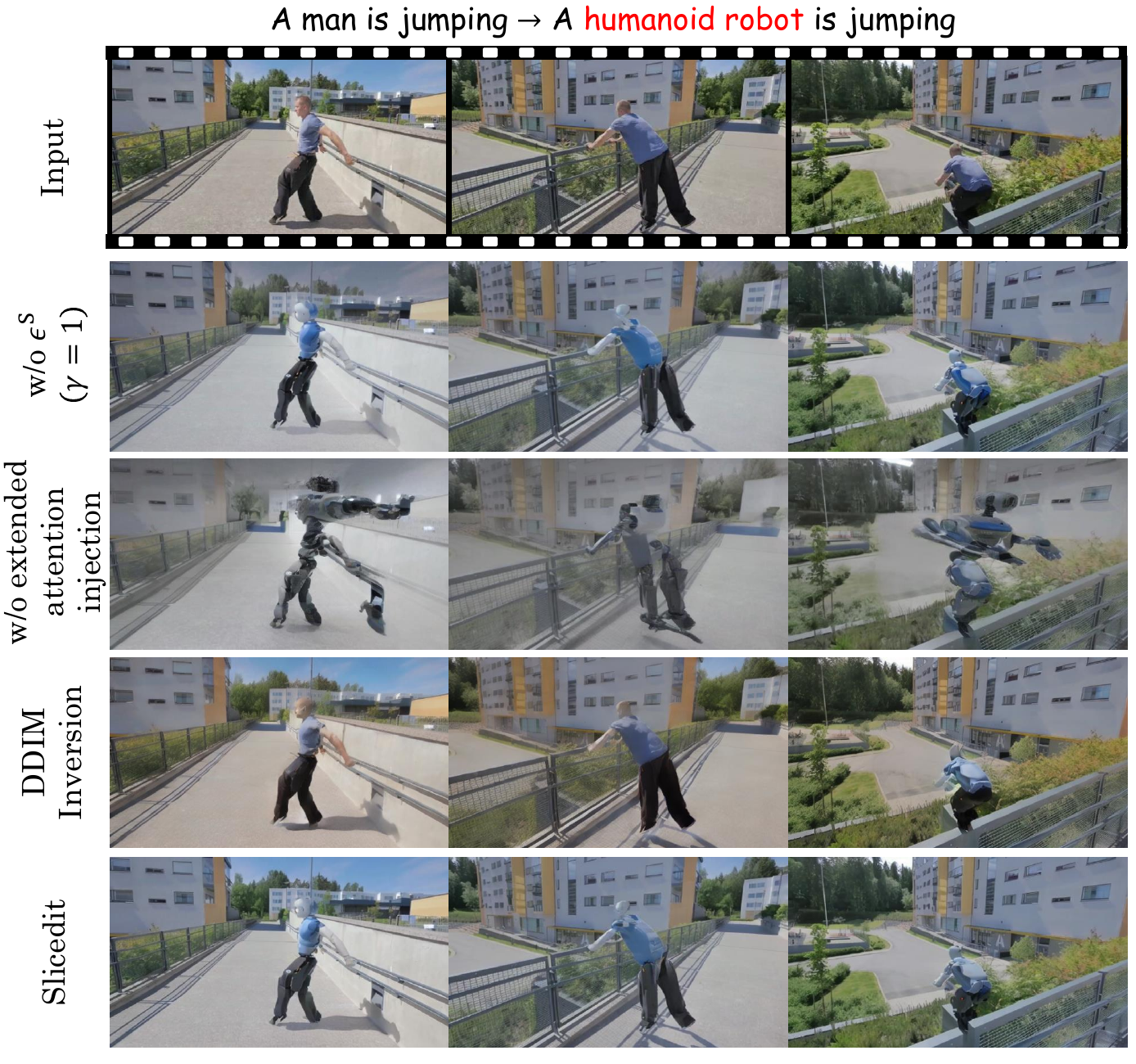}
\caption{\textbf{Ablation study.}}
\label{fig:SM_abaltion}
\end{figure}

Figure~\ref{fig:SM_abaltion} illustrates the importance of each component of our method. As can be seen in the frame comparison, and in a clearer way in the video comparison available on our \href{https://matankleiner.github.io/slicedit/\#ablation}{website}, without the spatiotemporal slices, the resulted video lacks temporal consistency. Specifically, each frame displays a different edit outcome, resulting in a jittery video. Without the extended attention injection the resulted video adheres to the text prompt but disregards the original video layout. When using DDIM inversion with the same parameters as used for DDPM inversion our method is not able to successfully edit the video according to the text prompt. For additional results with different configurations of DDIM inversion and more comparisons between DDPM and DDIM inversion see our \href{https://matankleiner.github.io/slicedit/supp.html#ddim}{supplemental website}.

\clearpage
\section{Hyperparameters}
\label{sm:hyperparams}
We evaluate the experiment (described in Sec.~\ref{sec:experiments}) with two different hyperparameters values of $T_\text{skip}$, strength, and the percentage of extended attention injection 

Table~\ref{tab:SM_hyperparams} provides a summary of the quantitative results. While the (8,14,85) configuration yields a slightly higher CLIP-Text score, the (8,10,85) configuration exhibits superior temporal consistency and minimizes LPIPS. We chose the latter configuration as the default for our method, employed to produce all the results in the paper, the Appendix, and the website. It is important to note that we did not explore different strengths for $\epsilon_{\theta}^\text{S}$ and $\epsilon_{\theta}^\text{EA}$.
Figure~\ref{fig:metrics} illustrates the results obtained with the varied hyperparameters, providing a comparison with the competing methods.

\begin{table}[h]
\centering
\caption{\textbf{Metrics over different hyperparameter set.} Table shows the metrics over a different hyperparameter set ($T_\text{skip}$, strength, injection).}
\label{tab:SM_hyperparams}
    \begin{tabular}{|l|cccc|} \hline 
          &  CLIP-Text $\uparrow$ & CLIP-Consistency & Flow Err. $\downarrow$ & LPIPS $\downarrow$  \\ \hline\hline 
         Original & --- & 0.982 & --- & --- \\ \hline 
         Ours $(8,14,85)$& 0.332 & 0.982 & 0.275 & 0.22\\ 
         \hline
          Ours $(8,10,85)$ & 0.329 & 0.982 & 0.252  & 0.159  \\ \hline   \end{tabular}

\end{table}

\clearpage

\section{Extended Attention}
\label{sm:ext_attent}
Self-attention~\citep{lin2017selfattention, vaswani2017attention} is attention mechanism that allows the model to relate to different parts of the same input. In images, self-attention, considers pixel location in feature maps to calculate similar correlation, for a given image. 

Given a video frame, $x^t$, or its latent representation, $z_{{x}^{t}}$, the self attention operation starts by projecting it to queries $Q$, keys $K$ and values $V$ of dimension $d$ by using learnable projection matrices, $W^Q, W^K, W^V$.
The attention is then computed as follows  
\begin{equation}
\label{eq:attention}
\text{Attention}(Q,K,V) = \text{Softmax}\left(\frac{QK^T}{\sqrt{d}}\right)\cdot V.
\end{equation}

This self-attention mechanism is suited for an image, or a single frame, but not for handling multiple frames together. Thus, an extended attention mechanism, a sparse version of causal attention, applying the attention mechanism to multiple frames was suggested by~\citet{Wu23TuneAVideo} for video editing. This attention mechanism or some version of it, dubbed in previous and concurrent works as spatiotemporal attention, cross-frame attention and extended attention (as we use in this work), is used in various image editing methods, ones that require tuning and zero-shot alike~\citep{Wu23TuneAVideo, QI2023FateZero, Ceylan2023Pix2Video, liu2023video-p2p, Wang2023vid2vid-zero, Yang23Rerender, geyer2024tokenflow, xing2023simda}

Extended attention enables the attention module to process multiple frames together, resulting in an attention map with correspondence between multiple frames. 
Given multiple video frames, $\{x^t,x^{t+a},x^{t+b},x^{t+c},...\}$ where $\{a,b,c,...\}$, are some scalars, or their latent representation $\{z_{{x}^{t}},z_{{x}^{t+a}},z_{{x}^{t+b}},z_{{x}^{t+c}},...\}$, the queries, keys and values are 
\begin{equation}
\label{eq:ext_att_qkv}
\begin{aligned}
Q &= W^Q \cdot z_{{x}^{t}}, \\
K^{E} &= W^K \cdot [z_{{x}^{t}},z_{{x}^{t+a}},z_{{x}^{t+b}},z_{{x}^{t+c}},...], \\
V^{E} &= W^V \cdot [z_{{x}^{t}},z_{{x}^{t+a}},z_{{x}^{t+b}},z_{{x}^{t+c}},...],
\end{aligned}
\end{equation}
where $[\cdot]$ denotes the concatenation operation.

Thus, the extended attention can be formulated as 
\begin{equation}
\label{eq:ext_attention}
\text{Extended-Attention}(Q,K^E,V^E) = \text{Softmax}\left(\frac{Q K^E}{\sqrt{d}}\right)\cdot V^E.    
\end{equation}

\begin{figure}[h]
\centering
\includegraphics[width=0.95\textwidth]{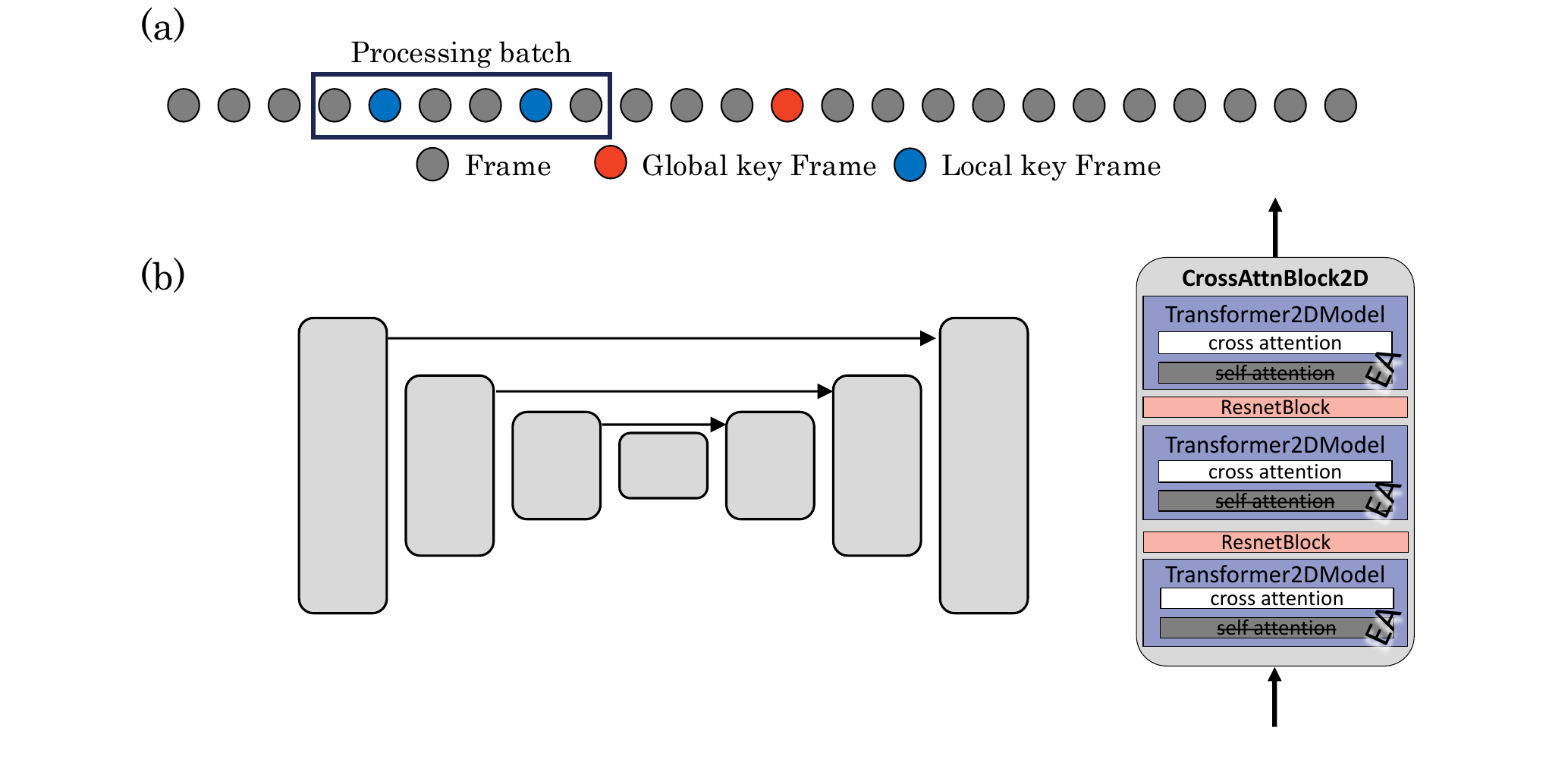}
\caption{\textbf{Extended attention.} (a) We calculate the extended attention between each frame (circle) within a processing batch (rectangle), and a set of three key-frames. In our implementation, the processing batch is 6. The key-frames are composed of a fixed global key-frame (red circles) and two local key-frames (blue circles). The global key-frame is located at half of the video duration. The Local key frames are the 2\textsuperscript{nd} and 5\textsuperscript{th} frames within the processing batch. (b) The U-Net network (left) is composed of a CrossAttnBlock2D (right) in each layer. As shown, all self attention layers are changed into extended attention (marked in gray).}
\label{fig:SM_ext_attn}
\end{figure}

Our extended attention is calculated between each frame and a set of three key-frames, consisting of one global frames positioned at half of the video length, along with two local frames which are chosen from within the processing batch. Our extended attention implementation is illustrated in Fig.~\ref{fig:SM_ext_attn}.

Note that while we implement extended attention in all transformer blocks, we use extended attention injection only in layers 4-11 in the upsample blocks of the U-Net, similarly to ~\citet{Tumanyan2023PnP}.

\clearpage

\section{Denoiser Experiment}
\label{sm:denoiser_exp}

We evaluated the performance of the Stable Diffusion denoiser across spatiotemporal slices in an experiment. For this experiment, we used 80 natural videos with different type of humans, animals and vehicles motion and of different environments, taken from \url{https://www.pexels.com/videos/}. Initially, we encoded each video frame by frame into the latent space of Stable Diffusion using a pretrained encoder. From these latent videos, we extracted frames and spatiotemporal slices ($\mathbf{y}-t$ slices). Additionally, we randomly permuted the pixels of each extracted frame to create a permuted frame. The permuted frames, which have noise-like patterns, serve as an out of distribution examplse to validate our experiment. In the latent space we added different levels of noise, following Stable Diffusion noise scheduler, to each frame, spatiotemporal slice and permuted frame. We then used Stable Diffusion pretrained denoiser to predict the added noise. We repeated this process across videos for 10000 samples and calculated the mean square error (MSE) between the predicted noise and the added noise. 

This experiment results, displayed in Fig.~\ref{fig:MSE_sd_denoiser}, illustrate that Stable Diffusion pretrained denoiser can successfully predict the noise added to spatiotemporal slices. It can be seen that for high noise levels the results for frames, spatiotemporal slices and permuted frames are almost the same. For lower level of noise, it can be seen that Stable Diffusion denoiser can successfully predict the noise added to spatiotemporal slices, and that it even predicts it better than it predicts the noise added to frames. It also can be seen that the pretrained denoiser struggles with predicting the noise added to the permuted frames, which are completely out of distribution examples. 

As we used the pretrained denoiser over the spatiotemporal slices only to induce smoothness along the temporal direction, its ability to successfully predict the added noise is all we need.

\clearpage

\section{Algorithm Pseudo Code}
\label{sm:algorithms}

In this section we provide pseudo code for editing a video using Slicedit. We first cover the pseudo code for DDPM inversion and editing using DDPM inversion, as presented by \citet{hubermanspiegelglas23}, as it is a key component of our video editing algorithm. Then we provide the pseudo code for Slicedit.  

\subsection{Editing using DDPM Inversion}
\label{sm:algorithmDDPM}
We note that this section is not a detailed explanation about editing using DDPM inversion, but a shorter one, relying on some previous knowledge about diffusion reverse and forward process. For a detailed discussion about DDPM inversion see~\citet{hubermanspiegelglas23}.  

Generating an image using diffusion process starts from a random noise vector, $x_T \sim \mathcal{N} (0, \mathbf{I})$ and iteratively denoises it using the recursion 
\begin{equation}
    x_{\tau-1}=\hat{\mu}_{\tau}(x_{\tau})+\sigma_{\tau}z_{\tau},\quad \tau=T,...,1
\end{equation}
where $\{z_{\tau}\}$ are iid standard normal vectors and 
\begin{equation}
\hat{\mu}_{\tau}(x_{\tau}, p) = \sqrt{\bar{\alpha}_{\tau - 1}}P(\epsilon_\theta(x_{\tau}, p)) + D(\epsilon_\theta(x_{\tau}, p)).    
\end{equation}
Where, $\epsilon_{\theta}(\cdot,p)$ is the denoiser, $p$ is a text prompt, $P(\epsilon_\theta(x_{\tau})) = (x_{\tau} - \sqrt{1 - \bar{\alpha_{\tau}}}\epsilon_\theta(x_{\tau}))/\sqrt{\bar{\alpha}_{\tau}}$ is the predicted $x_0$
and $D(\epsilon_\theta(x_{\tau})) = \sqrt{1 - \bar{\alpha}_{\tau-1}-{\sigma_{\tau}}^2}\epsilon_\theta(x_{\tau})$ is a direction pointing to $x_{\tau}$. 

In order to edit a real signal, $x_0$, it first needs to be inverted into a diffusion model noise space, meaning to extract the $\{x_T, z_T, ..., z_1\}$ that produced this signal. These noise vectors are in the latent space, means their dimensions are $64\times64\times4$. \citet{hubermanspiegelglas23} proposed the edit friendly DDPM noise space, which can be extracted using Alg.~\ref{alg:ddpm-inv}. In Alg.~\ref{alg:ddpm-inv} the denoiser is conditioned on a text prompt describing the real signal, $p_{\text{src}}$, however, the noise space can be extracted without text conditioning.   

\begin{algorithm}[h]
\begin{algorithmic} 
\caption{Edit friendly DDPM inversion}\label{alg:ddpm-inv}
\STATE {\bfseries Input :} $x_0, p_{\text{src}}$
\STATE {\bfseries Output} : $\{x_T, z_T, ..., z_1\}$
\FOR{$\tau = 1$ \textbf{to} $T$}
\STATE $\tilde{\epsilon} \sim \mathcal{N}(0,\,1) $
\STATE $x_{\tau} \leftarrow \sqrt{\bar{\alpha_{\tau}}}x_0 + \sqrt{1 - \bar{\alpha_{\tau}}}\tilde{\epsilon}$
\ENDFOR 
\FOR{$\tau = 1$ \textbf{to} $T$}
\STATE $z_{\tau} \leftarrow (x_{\tau-1} - \hat{\mu}_{\tau}(x_{\tau}, p_{\text{src}}))/\sigma_{\tau}$ \hfill\texttt{{//shape } $64\times64\times4$}

\ENDFOR
\STATE {\bfseries return} $\{x_T, z_T, ..., z_1\}$
\end{algorithmic}
\end{algorithm}

After extracting the noise space, the signal can be edited by applying the diffusion process and condition the denoiser on a target prompt, describing the desired edit, as summarized in Alg.~\ref{alg:edit}.  

\begin{algorithm}[h]
\caption{Editing}\label{alg:edit}
\begin{algorithmic}
\STATE {\bfseries Input :} $\{x_T, z_T, ..., z_1, p_{\text{tar}}\}$
\STATE {\bfseries Output} : $\tilde{x}_0$
\STATE $\tilde{x}_T \leftarrow x_T$
\FOR{$\tau = T$ \textbf{to} $1$}
\STATE $\tilde{x}_{\tau-1} \leftarrow  \hat{\mu}_{\tau}(\tilde{x}_{\tau}, p_{\text{tar}}) + \sigma_{\tau}z_{\tau}$
\ENDFOR
\STATE {\bfseries return} $\tilde{x}_0$

\end{algorithmic}
\end{algorithm}

\clearpage

\subsection{Slicedit}
\label{sm:algorithmVideo}
Slicedit follows a similar structure, mainly first inverting the input video, $I_0$, consisting $N$ frames, into edit friendly noise space and then applying the diffusion process, while conditioning the denoiser on a text prompt. 
A key difference is the denoiser itself, as in Slicedit the denoiser is $\epsilon^\text{V}_{\theta}(\cdot,p)$, which is described in detail in Sec.~\ref{sec:method}. 

The resulting algorithm, Alg.~\ref{alg:slicedit}, given with the notations from the main paper is as follows

\begin{algorithm}[H]
\caption{Slicedit video editing}\label{alg:slicedit}
\begin{algorithmic}
\STATE {\bfseries Input :} $I_0, p_{\text{src}}, p_{\text{tar}}$
\STATE {\bfseries Output :} $J_0$
\STATE \texttt{{// }Video inversion}
\FOR{$\tau = 1$ \textbf{to} $T$}
\STATE $\tilde{\epsilon} \sim \mathcal{N}(0,\,1) $
\STATE $I_\tau \leftarrow \sqrt{\bar{\alpha_\tau}}I_0 + \sqrt{1 - \bar{\alpha_\tau}}\tilde{\epsilon}$
\ENDFOR 
\FOR{$\tau = 1$ \textbf{to} $T$}
\STATE $Z_\tau \leftarrow (I_{\tau-1} - \hat{\mu}_\tau^V(I_\tau, p_{\text{src}}))/\sigma_\tau$ \hfill\texttt{{//shape } $N\times 64\times64\times4$}
\STATE $Q_\tau, K_\tau \leftarrow \epsilon^\text{V}_{\theta}(\cdot,p_{\text{src}})$  \hfill\texttt{{//} Keep keys and queries of the input video}
\ENDFOR
\STATE \texttt{{// }Video editing}
\STATE $J_T \leftarrow I_T$
\FOR{$\tau = T$ \textbf{to} $1$}
\STATE $\epsilon^\text{V}_{\theta}(\cdot,p_{\text{tar}}) \leftarrow Q_\tau, K_\tau$ \hfill\texttt{{//} Inject input video keys and queries to the edited video}
\STATE $J_{\tau-1} \leftarrow  \hat{\mu}_{\tau}^V(J_{\tau}, p_{\text{tar}}) + \sigma_{\tau}Z_{\tau}$
\ENDFOR

\STATE {\bfseries return} $J_0$
\end{algorithmic}
\end{algorithm}

where $\hat{\mu}_{\tau}^\text{V}(\cdot, p) = \sqrt{\bar{\alpha}_{\tau - 1}}P(\epsilon_\theta^\text{V}(\cdot, p)) + D(\epsilon_\theta^\text{V}(\cdot, p))$ and $\epsilon_\theta^\text{V}(\cdot, p)$ is defined in Eq.~\ref{eq:epsilon_volume}.

\end{document}